\documentclass[letterpaper]{article} 
\usepackage{aaai25}  
\usepackage{times}  
\usepackage{helvet}  
\usepackage{courier}  
\usepackage[hyphens]{url}  
\usepackage{graphicx} 
\usepackage{natbib}  
\usepackage{caption} 
\usepackage{algorithm}
\usepackage{algorithmic}

\usepackage{newfloat}
\usepackage{listings}
\DeclareCaptionStyle{ruled}{labelfont=normalfont,labelsep=colon,strut=off} 
\floatstyle{ruled}
\newfloat{listing}{tb}{lst}{}
\floatname{listing}{Listing}
\usepackage{amsfonts}
\usepackage{amsmath}
\usepackage{amssymb}
\usepackage{mathrsfs}
\usepackage{booktabs} 
\usepackage{multirow} 
\usepackage{xcolor}
\usepackage{float}
\usepackage{graphicx}
\usepackage{subfig}
\usepackage{threeparttable}

\newtheorem{theorem}{Theorem}

\newtheorem{lemma}{Lemma}
\newtheorem{definition}{Definition}
\newtheorem{example}{Example}
\newtheorem{assumption}{Assumption}
\newcommand{\M}{LoGoFair}
\title{LoGoFair: Post-Processing for Local and Global Fairness in Federated Learning}
\author {
    Li Zhang\textsuperscript{\rm},
    Chaochao Chen\textsuperscript{\rm}\thanks{Chaochao Chen is the corresponding author.},
    Zhongxuan Han\textsuperscript{\rm},
    Qiyong Zhong\textsuperscript{\rm},
    Xiaolin Zheng\textsuperscript{\rm}
}
\affiliations {
    \textsuperscript{\rm}Zhejiang University\\
    zhanglizl80@gmail.com, \{zjuccc, zxhan, youngzhong, xlzheng\}@zju.edu.cn
}

\begin{document}

\maketitle
\begin{abstract}
Federated learning (FL) has garnered considerable interest for its capability to learn from decentralized data sources.
Given the increasing application of FL in decision-making scenarios, addressing fairness issues across different sensitive groups (e.g., female, male) in FL is crucial.
Current research often focuses on facilitating fairness at each client's data (\textit{local fairness}) or within the entire dataset across all clients (\textit{global fairness}).
However, existing approaches that focus exclusively on either local or global fairness fail to address two key challenges: 
(\textbf{CH1}) \textit{Under statistical heterogeneity, global fairness does not imply local fairness, and vice versa.}
(\textbf{CH2}) \textit{Achieving fairness under model-agnostic setting.}
To tackle the aforementioned challenges, this paper proposes a novel post-processing framework for achieving both \textbf{Lo}cal and \textbf{G}l\textbf{o}bal \textbf{Fair}ness in the FL context, namely \M.
\textit{To address CH1}, \M~endeavors to seek the Bayes optimal classifier under local and global fairness constraints, which strikes the optimal accuracy-fairness balance in the probabilistic sense.
\textit{To address CH2}, \M~employs a model-agnostic federated post-processing procedure that enables clients to collaboratively optimize global fairness while ensuring local fairness, thereby achieving the optimal fair classifier within FL.
Experimental results on three real-world datasets further illustrate the effectiveness of the proposed \M~framework.
Code is available at~\url{https://github.com/liizhang/LoGofair}.
\end{abstract}

\section{Introduction}
Federated learning is a distributed machine learning paradigm that enables multiple clients to collaboratively refine a shared model while preserving their data privacy~\cite{fedavg}. 
With the growing integration of FL in high-stakes scenarios such as healthcare~\cite{healthcare-2,fedllm}, finance~\cite{finance-chouldechova2017fair}, and recommendation systems~\cite{recommendation-burke2017multisided}, fairness is gaining prominence to prevent machine learning models from discriminating any demographic group based on sensitive attributes, e.g. gender and race. 
Several methods exist to achieve group fairness in centralized settings~\cite{reduction-agarwal2018reductions,beyond-alghamdi2022beyond,pre-processing-jovanovic2023fare,posthoc}, these methods typically require direct access to entire datasets, thereby incurring high communication costs and privacy concerns if directly implemented in the FL environment.

\begin{figure}[t!]
\vspace{-15pt}
   \centering
   \captionsetup{font=small, labelfont=small}
    \subfloat[Local/Global Fairness]{
	\label{introduction-Local/Global-Fairness}
	\includegraphics[width=0.23\textwidth]{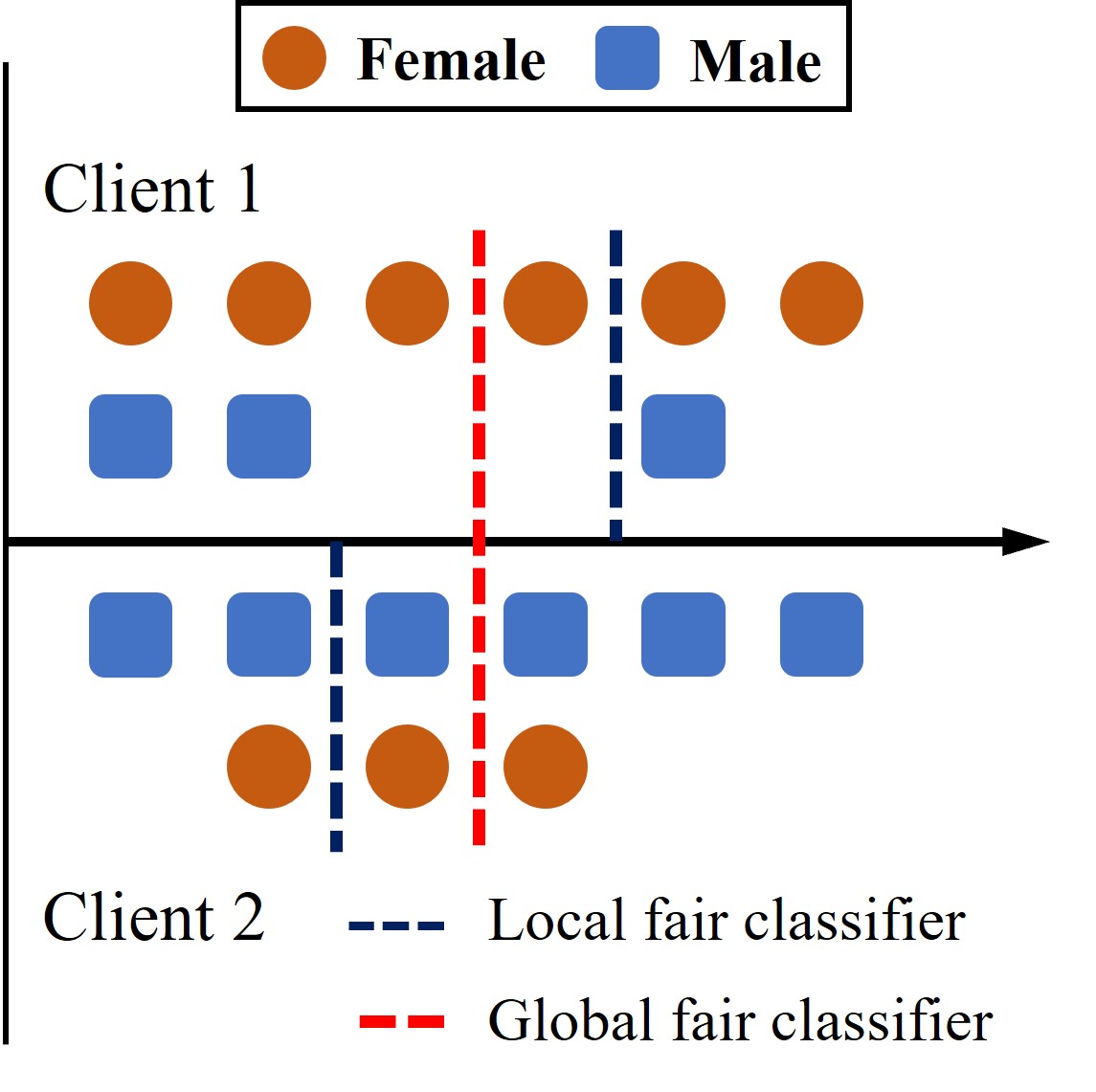}}
    \subfloat[Fairness under Heterogeneity]{
	\label{introduction-Heterogeneity}
	\includegraphics[width=0.23\textwidth]{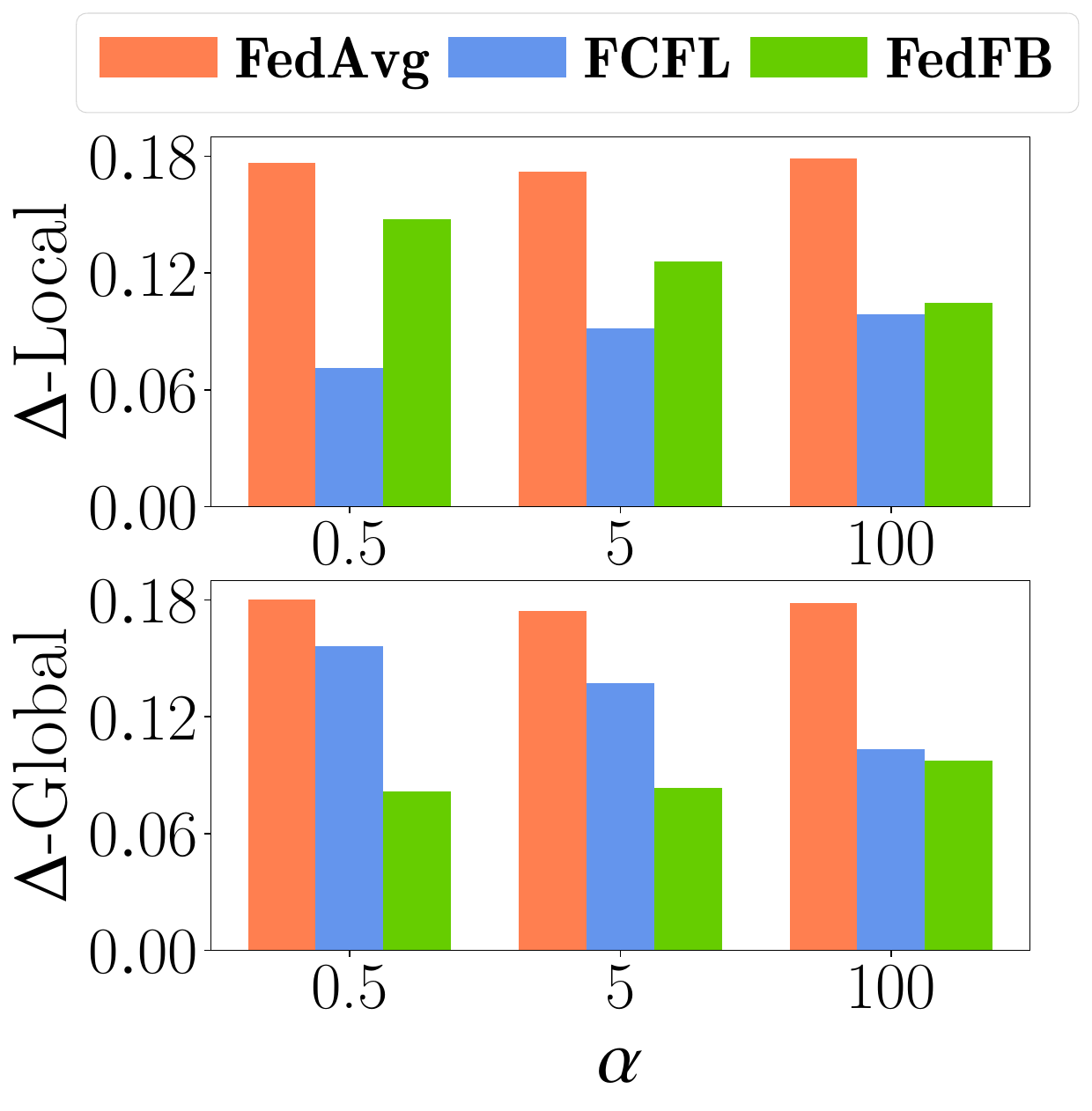}}
    \caption{\small{(a) Demonstrates a toy example of \textit{local and global fairness} in the context of an one-dimensional classification problem. These fair classifiers ensure equal gender proportions in classification at local and global level (e.g. Local classifiers allocate 2/3 of samples from both genders to the left side for client 1). (b) Presents comparisons between local (FCFL) and
global (FedFB) fair FL algorithms across varying levels of heterogeneity. A smaller $\alpha$
signifies more heterogeneity across clients, and a smaller $\Delta$ denotes a fairer model at local or global level.}}
\vspace{-15pt}
\end{figure}
%

%

%
%

%
To develop fairness guarantees for federated algorithms, this paper focuses on two key concepts of group fairness in FL: Local and Global Fairness~\cite{FCFL-cui2021addressing,fairfed,AGLFOP-hamman2024demystifying}.
\textbf{Local fairness} aims to develop models that deliver unbiased predictions across specific groups when evaluated on each client's local dataset.
%
%
Since the models are ultimately deployed and applied in local environments, achieving local fairness is indispensable for promoting fair FL models.
\textbf{Global fairness} focuses on identifying models that ensure similar treatment for sensitive groups within the entire dataset across all clients.
In practice, models trained on large-scale aggregated datasets are inclined to learn inherent bias in data and exacerbate the treatment discrepancy of sensitive groups based as shortcuts to achieving high accuracy.~\cite{shortcut-geirhos2020shortcut,bias-propagation-chang2023bias}.
%
These global models typically fail to make impartial decisions and uphold societal fairness.
Figure~\ref{introduction-Local/Global-Fairness} provides an example of local and global fairness, illustrating that these two fairness notions can differ.
Therefore, both local and global fairness are crucial for achieving group fairness in the FL setting. 
Recent works have introduced several techniques for achieving either local or global fairness. Most of them enhance fairness through dynamically adapting aggregation weights~\cite{FCFL-cui2021addressing,agnostic-FL-du2021fairness,fairfed} or reweighting the training samples~\cite{mitigate-bias-abay2020mitigatingbiasfederatedlearning,fedfb-zeng2021improving}.
However, existing methods face certain challenges~(CH) in achieving both local and global fairness. 
\textbf{CH1}: \textit{Under statistical heterogeneity, local fairness does not imply global fairness, and vice versa.}
(1)~Statistical heterogeneity in FL leads to varying representations of sensitive groups across clients, causing local biases to differ from those in the globally aggregated dataset~\cite{FCFL-cui2021addressing}.
As a result, methods that lessen global bias are unable to specifically tackle local bias and ensure fairness at the client level, similar to methods focused on local fairness.
Pervious work~\cite{AGLFOP-hamman2024demystifying} rigorously proved that neither local nor global fairness implies the other in FL, and further empirically examining this result across various cases of heterogeneity.
(2)~It is uncertain whether the local model will retain its debiasing capability after model aggregation, especially for nonlinear deep learning models~\cite{bias-propagation-chang2023bias}.
%
Figure~\ref{introduction-Heterogeneity} compares the fairness performance of methods tailored for either local or global fairness.  
The effectiveness of the comparison algorithms in the other fairness notion present a dramatic decline with increasing data heterogeneity.
Hence, it is explicitly impracticable to achieve both local and global fairness by focusing solely on one of them within FL.
\textbf{CH2:} \textit{Achieving fairness under model-agnostic setting}.
Most existing approaches employ the \textit{in-processing} strategy~\cite{FCFL-cui2021addressing,agnosticFL-du2021fairness,fairfed}, which entails intervening in the model training process and typically complicates the FL models.
However, retraining the federated model is often impractical due to significant communication overhead and privacy concerns.
An effective solution lies in investigating model-agnostic approaches to ensuring fairness in FL.
In this paper, we propose a novel federated post-processing framework for achieving both local and global fairness, namely~\M.
To the best of our knowledge, we are the first to offer both local and global fairness guarantees in the FL context.
%
%
Generally, \M~facilitates clients in collaboratively calibrating the pre-trained classifier under specific procedure crafted to guarantee both fairness notions.
To tackle \textbf{CH1}, we establish a specific characterization of the Bayesian optimal classifier under local and global fairness constraints.
%
%
\M~aims to achieve both fairness notions with minimal accuracy decline through learning this classifier.
%
%
To tackle \textbf{CH2}, \M~introduces a federated post-processing procedure that learn to calibrate federated model to the optimal fair classifier in a model-agnostic manner.
This approach enables participating clients to collaboratively enhance global fairness while reinforcing their individual local fairness. 
%
%
%
The numerical results demonstrate that \M~outperforms existing methods in achieving superior local and global fairness with competitive model accuracy.
Moreover, experiments also illustrate that \M~enables the flexible adjustment of the accuracy-fairness trade-off in the FL environment.

\noindent Our main contributions can be summarized as follows:
\begin{itemize}
    \item We propose a novel post-processing framework named \M~to achieve both local and global fairness in FL.
    \item We characterize an explicit formula of the Bayesian optimal classifier under local and global fairness constraints as the learning target of~\M. 
    \item We propose a federated post-processing procedure for clients to jointly optimize global fairness while guaranteeing local fairness, aiming to achieve the optimal fair classifier.
    \item Experimental results demonstrate that \M~outperforms existing methods in achieving superior balances among accuracy, local fairness, and global fairness. 
\end{itemize}

\section{Related work}
%
\subsection{Group Fairness in Machine Learning}
Group fairness in machine learning has grown rapidly over the last few years into a key area of trustworthy AI, especially in high-stakes decision-making systems, such as healthcare~\citep{healthcare-ml,fairml}, criminal prediction~\cite{Fairness-in-Criminal} and recommendation systems~\cite{User-oriented,User-Oriented-1,User-Oriented-2,user-oriented-3}.
As summarized in previous work~\cite{fair-survey-mehrabi2021survey}, group fairness is broadly defined as the absence of prejudice or favoritism toward a sensitive group based on their inherent characteristics.
In conventional centralized machine learning, common strategies for realizing group fairness can be classified into three categories: pre-processing, in-processing, and post-processing techniques.
\textbf{Pre-processing}~\cite{fair-pre-pmlr-v162-li22p,fair-pre-2-pmlr-v202-jovanovic23a,fair-pre-3-10.1145/3394486.3403080} approaches aim to modify training data to eradicate underlying bias before model training.
\textbf{In-processing}~\cite{inpro-pmlr-v162-kim22b,inpro-pmlr-v202-li23h} methods are developed to achieve fairness requirements by intervention during the training process.
\textbf{Post-processing}~\cite{fair-regression-chzhen2020fair,posthoc,fair-guarantees-denis2024fairness} methods adjust the prediction results generated by a given model to adapt to fairness constraints after the model is trained.
In this paper, we propose a post-processing technique specifically tailored to ensure group fairness in FL environments.
\subsection{Fair Federated Learning}
Fairness faces new challenges in FL context. 
Several recent works have introduced and investigated fairness concepts specific to FL, such as client-based fairness, and collaborative fairness. 
Nonetheless, the impact of FL on group fairness has yet to be thoroughly explored.
Existing methods primarily utilize in-processing strategies to address global or local fairness issues.
%

%
%

%
Concerning local fairness, prior work \cite{bias-propagation-chang2023bias} highlights potential detrimental effects of FL on the group fairness of individuals, while others \cite{FCFL-cui2021addressing} propose algorithms to enhance local fairness without compromising performance consistency.
%
%
%
Concerning global fairness, two main approaches are adaptive reweighting techniques~\cite{AFL-mohri2019agnostic,fairfed,fedfb-zeng2021improving,mitigate-bias-abay2020mitigatingbiasfederatedlearning} and solving federated optimization objectives with relaxed differentiable fairness constraints or regularization~\cite{agnostic-FL-du2021fairness,wang2023mitigating,Lag-dunda2024fairnessawarefederatedminimaxoptimization}.
(1) Reweighting techniques dynamically reweight clients or data during training time.
(2) Optimization methods solve objectives with relaxed fairness constraints or regularization.
%
%

%
Besides, previous work~\cite{AGLFOP-hamman2024demystifying} theoretically investigated the interplay between local and global fairness in FL.
They introduced AGLFOP to explore the theoretical limits of the accuracy-fairness trade-off, aiming to identify the optimal performance achievable given global data distributions and prediction outcomes.
Challenges that persist include limited flexibility and generalization ability in addressing unfairness issues in prediction tasks within intricate FL scenarios.
%
In this paper, we focus on both local and global fairness. \M~effectively achieves both fairness notions with flexible trade-offs.

\section{Preliminaries}

\subsection{Fairness Notion}
Let $(X, A, Y )$ be a random tuple, where $X \in \mathcal{X}$ for some feature space $\mathcal{X} \in \mathbb{R}^d$, labels $Y \in \mathcal{Y} = \{0, 1\}$ for a binary classification problem, and the sensitive attribute $A \in \mathcal{A}$. In machine learning, the concept of group fairness aims to ensure that ML models provide equitable treatment to individuals with diverse sensitive attributes, such as gender, race, and age.
Without loss of generality, this paper concentrates on two sensitive groups following~\cite{fedfb-zeng2021improving,fairfed}, with $\mathcal{A} = \{-1, 1\}$ representing the set of protected sensitive attributes. 
The goal of fair classification is to identify a attribute-aware classifier $h(x,a): \mathcal{X} \times \mathcal{A} \rightarrow \mathcal{Y}$ subject to the constraints imposed by the specified fairness criteria.
In this paper, we mainly focus on two popular group fairness criteria:
\begin{itemize}
    \item \textbf{Demographic Parity (DP)}~\cite{DP-dwork2012fairness} emphasizes that the positive rate of predictor $\widehat{Y}=h(X,A)$ is equal in each sensitive group, measured by
    \begin{equation*}
    \small
        \mathcal{M}_{DP}=\left | P(\widehat{Y} = 1 | A = -1) - P(\widehat{Y} = 1 | A = 1) \right |.
    \end{equation*}
    \item \textbf{Equalized Odds (EO)}~\cite{EO-hardt2016equality} concentrates on equalizing the false positive and true positive rates in each sensitive group, measured by
    \begin{align*}
         \mathcal{M}_{EO}=\underset{y \in \{0,1\}}{\max}\bigg| P(\widehat{Y}  = & 1 | A = -1, Y = y) -\\
         & P(\widehat{Y} = 1 | A = 1, Y = y) \bigg|.\\
    \end{align*}
\end{itemize}
%
%
%

%

%
%

\subsection{Fairness in Federated learning}
A federated system consists of numerous decentralized clients, so that we consider the global data space as $\mathcal{S} = \{\mathcal{X}, \mathcal{A}, \mathcal{Y}, \mathcal{C}\}$, where $\mathcal{C}$ represents the client index set with $|\mathcal{C}|$ denoting the number of total clients.
%
The global data distribution can be formally represented by jointly random variable $S=(X,A,Y,C) \in \mathcal{S}$.
%
%
FL focuses on the scenario in which the data is dispersed across $|\mathcal{C}|$ different clients, with each client $c$ processing a local data dataset $\mathcal{D}_c$, $c \in [|\mathcal{C}|]$. 
%
Each sample in $\mathcal{D}_c$ is assumed to be drawn from local distribution, represented as $(x_{c,i}, a_{c,i}, y_{c,i})$, where $i \in [N_c]$, and $N_c$ represents the number of samples for client $c$.
Since local fairness act on clients' individual data distributions, we use a attribute-aware fair classifier $h(x,a,c)$ for each client to modify federated model.
Here we introduce Bayes score function to characterize fairness notions $\eta(x,a):=P \left(Y=1|X=x, A=a\right)$, which possesses a natural extension in FL:
\begin{equation}
\label{bayes-score}
\eta(x,a,c):=P \left(Y=1 \mid X=x, A=a,C=c \right).
\end{equation}
%
%
%
%
FL aims to minimize global risk $\mathcal{R}(h) := P(h(X,A,C) \neq Y)$ through minimizing the weighted average of the loss across all clients: $\min_{\theta} L(\theta)=\sum_{c=1}^{|\mathcal{C}|} a_c L_c(\theta)$, and the local objective $L_c(\theta)=\frac{1}{N_c}\sum_{(x,y)\in \mathcal{D}_c}\ell(f(x;\theta),y)$, where $a_c$ signifies the importance coefficient; $\ell(\cdot,\cdot)$ denotes the loss function; $f$ indicates the federated model.
In this paper, we are interested in both local and global fairness criteria in the FL context following~\cite{fairfed,AGLFOP-hamman2024demystifying}. 
%

\begin{definition}
\label{local-def}
    {\rm\textbf{(Local Fairness and Global Fairness)}} The local fairness measures the disparity regarding sensitive groups aroused by the federated model when evaluated on the individual data distribution of each client, while the global fairness measures the disparity regarding sensitive groups on global data distribution across all clients.
\end{definition}
%
%
%
%
Formally, local data distribution can be represented by conditional distribution $P(X,A,Y| C)$, while global federated distribution is $P(X,A,Y)$.
The disparity on the treatment of sensitive groups can be qualified by the fairness metrics ($\mathcal{M}_{DP}$ and $\mathcal{M}_{EO}$) introduced before.
%
%


\section{LoGoFair}
\subsection{Overview}
\begin{figure}[t]
   \centering
	\includegraphics[width=\linewidth]{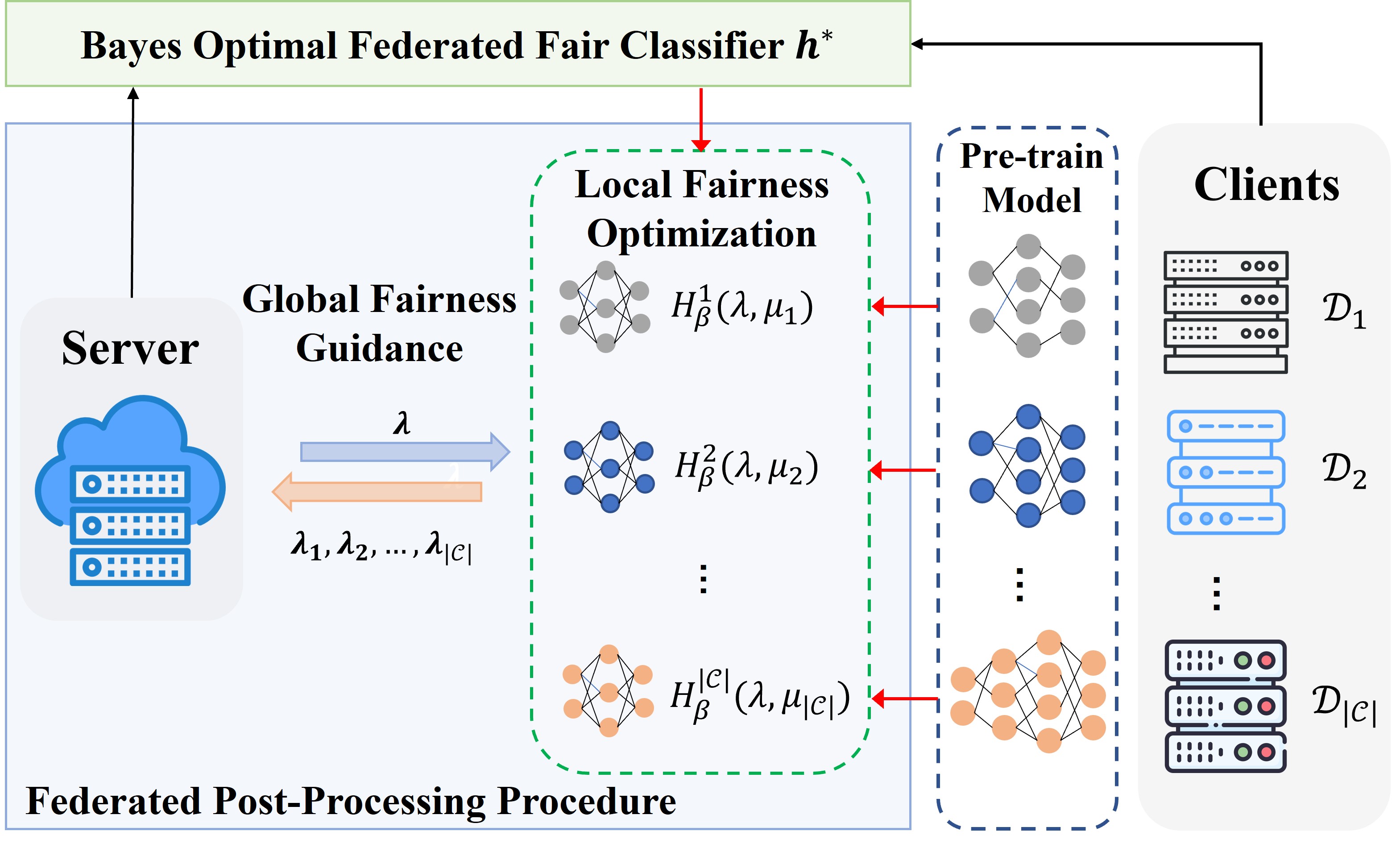}
    \captionsetup{font=small, labelfont=small}
    \caption{\small Overview of \M~framework. The \textit{Bayes optimal federated fair classifier}, which strikes the optimal accuracy-fairness balance, is identified as the objective. The \textit{federated post-processing procedure} reformulates it as a bi-level problem incorporating local fairness optimization and global fairness guidance.} 
    \label{model-framework}
    \vspace{-15pt}
\end{figure}

In this paper, we propose a novel post-processing framework to achieve both local and global fairness, namely~\M. 
The goal of \M~is to guarantee fairness in the FL setting with minimal accuracy sacrifice, while remaining compatible with either local or global fairness.
Designed as a post-processing framework, \M~can be conveniently adopted to mitigate bias in a wide range of existing FL algorithms.
As illustrated in Figure~\ref{model-framework}, our proposed method comprises two primary phases in the post-processing framework.
In the first phase, \M~deduces the representation of the Bayes optimal federated fair classifier that realizes both local and global fairness with maximal accuracy from aggregated information. 
This classifier can be achieved by optimizing a convex minimization problem, as proved in Theorem~\ref{theo-1-the-optimal}.
In the second phase, the optimization regarding the fair classifier is reformulated as a bilevel problem after the training stage, incorporating local fairness optimization and parameterized global fairness guidance shared across all clients.
Inspired by this problem formulation, \M~establishes a federated post-processing procedure that enables clients to collaboratively enhance global fairness while refining their local fairness.
The next two sections will investigate these two phases in detail.

\subsection{Bayes optimal federated fair classifier}
As demonstrated in \textbf{CH1} and Figure~\ref{introduction-Heterogeneity}, models that focus solely on global or local fairness tend to fail in the other metric under statistical heterogeneity. 
This phenomenon compels us to account for fairness constraints at the both levels when developing fair FL models.
To prevent excessive accuracy degradation due to fairness issues, we endeavor to solve the essential problem of what is the optimal classifier when local and global fairness restrictions are imposed.
In this section, we derive the optimal fair classifier from the Bayes optimal score function as the learning target of~\M.
We are interesting in either DP and EO metrics in this paper. 
%
%
For the sake of generality, we perform our analysis on \textit{composite linear disparity}, a specific fairness metric that characterizes the group-wise disparity with Bayes optimal score.
For a given distribution $\mathcal{P} \sim (X,A,Y)$, denote the marginal distribution of $X$ by $\mathcal{P}^X$, the conditional distribution of $\mathcal{P}^X$ on sensitive group $A=a$ by $\mathcal{P}^X_{a}$.
Given a classifier model $h(x,a)$, the \textit{composite linear disparity} of $h$ over $\mathcal{P}$ is defined as $\mathcal{M}_{\phi} (h) = \max_{k=1,\ldots,K} \left| \mathscr{D}_k (h) \right|$, and
\begin{equation*}
    \begin{split}
    \mathscr{D}_k (h) = \sum_{a\in \mathcal{A}} \underset{X \sim \mathcal{P}^X_{a}}{\mathbb{E}}\left[ \phi_k^a ( \eta(X,a)) h(X,a) \right].
    \end{split}
\end{equation*}
where $\{\phi_k^a\}_{k=1}^K$ are linear functions that depend on $a$, and $K$ is the number of linear disparities required to characterize the fairness criterion. 
This fairness metric encompasses DP and EO, along with other fairness notions such as Equality of Opportunity~\cite{EO-hardt2016equality} and Predictive Equality~\cite{PE-chouldechova2017fair}.
%
%
We give examples here to specify the fairness metrics represented by this notion.
\begin{example}
\label{exmple-1}
    {\rm(DP and EO).} We have a sensitive attribute $a \in \mathcal{A} = \{-1, 1\}$.\\
    {\rm1)} Let $K=1$ and $\phi_1^a=a$, we get DP metric. \\
    {\rm2)} Let $K=2$ and $\phi_1^a(w)=aw/P(Y=1 | A=a)$, $\phi_2^a(w)=a(1-w)/P(Y=0 | A=a)$, we get EO metric.
\end{example}

Furthermore, we extend \textit{composite linear disparity} to the local and global fairness notions based on Definition~\ref{local-def}.
With the marginal distribution of $X$ given sensitive attribute $a$ and client $c$ denoted by $\mathcal{P}^X_{a,c}$, the local and global fairness can be represented by 
\begin{equation*}
\small
    \begin{split}
    {\rm(Local) \ }&: \quad \mathcal{M}^{l,c}_{\psi} (h)=\max_{k_{l}=1,\ldots,K_{l,c}} \left| \mathscr{D}^{l,c}_{k_{l}} (h) \right|, \\
    \mathscr{D}^{l,c}_{k_{l}}(h)&=\sum_{a \in \mathcal{A}} \underset{X\sim \mathcal{P}^X_{a,c}}{\mathbb{E}}\left[ \psi^{a,c}_{k_{l}}( \eta(X,a,c) ) h(X,a,c) \right],\\
    {\rm(Global)}&: \quad \mathcal{M}^g_{\phi} (h)=\max_{k_g=1,\ldots,K_g} \left| \mathscr{D}^g_{k_g} (h) \right|,\\
    \mathscr{D}^g_{k_g}(h)&=\sum_{a \in \mathcal{A}} \sum_{c\in\mathcal{C}} \underset{X\sim \mathcal{P}^X_{a,c}}{\mathbb{E}}\left[ \phi_{k_g}^{a,c} ( \eta(X,a,c) ) h(X,a,c) \right],
    \end{split}
\end{equation*}
where $\{\psi_{k_l}^{a,c}\}_{k_l=1}^{K_{l,c}}$ and $\{\phi_{k_g}^{a,c}\}_{k_g=1}^{K_g}$ are the linear functions determined by $a$ and $c$, and $(K_{l,c},K_g)$ are the number of local and global linear disparities. 
The detailed proof and more examples within FL are provided in \textbf{Appendix A}.
To investigate the optimal classifier with fairness guarantee within FL, we consider the situation that there is a unified fairness constraint in global level, and each client has additional local fairness restrictions in response to personalized demands.
This problem can be interpreted as: \textit{which classifier $h$ minimizes the misclassification risk $\mathcal{R}(h)$ under the restriction that the local and global fairness measures $(\mathcal{M}_{\psi}^{l,c} (h),\mathcal{M}_{\phi}^g (h))$ are below given positive levels $(\delta^l,\delta^g)$.}
%
%
Denoting the set of all classifiers by $\mathcal{H}$, the problem can be formulated as
\begin{equation}
{\small
    \label{var-problem}}
    \begin{split}
     &\min _{h \in \mathcal{H}} \mathcal{R}(h),\\
     \text { s.t. } & |\mathscr{D}^{l,c}_{k_l}(h)| \leq \delta^{l,c}, c \in [C],k_l \in [K_{l,c}],\\
     & |\mathscr{D}^{g}_{k_g}(h)| \leq \delta^{g}, k_g \in [K_g]. \\
    \end{split}
\end{equation}
The solution of Problem~\eqref{var-problem} is identified as the \textit{Bayes optimal federated fair classifier}, which theoretically achieves the optimal trade-off among accuracy, global fairness and local fairness in the FL context. 
In order to establishing the closed form of optimal fair classifier, we introduce following mild assumption.
%
\begin{assumption}
    \label{assumpt-1}
    {\rm($\eta$-continuity).} For each client $c$, we assume that the mappings $t \mapsto P(\eta(X, A, C) \leq t \mid A=a,C=c)$ are almost surely continuous restricted to $[0,1]$.
\end{assumption}
Assumption~\ref{assumpt-1} implies that the conditional probability function, when regarded as a random variable, has no atoms.
It can be fulfilled from any estimator $\eta(x,a,c)$ and conditional distribution $\mathcal{P}^X_{a,c}$ by adding slight random noises~\cite{leveraging-chzhen2019leveraging,fair-regression-chzhen2020fair,fair-guarantees-denis2024fairness}. 
In brief, this assumption is of little or no concern in practice, as demonstrated in our experiments in section 5.
In the following result, we verify that the optimal federated fair classifier can be attained by calibrating the Bayes optimal score function.
\begin{theorem}
    \label{theo-1-the-optimal}
    {\rm (Bayes optimal federated fair classifier).}
    %
    Under Assumption~\ref{assumpt-1}, for $p_{a,c} := P(A=a,C=c)$, $\phi^{a,c}:=[ \phi^{a,c}_{k_g} ]_{k_g=1}^{K_g}$ and $\psi^{a,c}:=[ \phi^{a,c}_{k_l} ]_{k_l=1}^{K_{l,c}}$, $h^*$ is the Beyes optimal federated fair classifier which solves problem~\eqref{var-problem} if 
    \begin{flalign*}
    \small
    h^*(x,a,c) = \mathbb{I} \left[F(\lambda^*,\mu^*,x,a,c) \geq 0\right],
    \end{flalign*}
    and calibration function
    \begin{equation}
    \small
    \label{F}
    \begin{split}
    F(\lambda^*&,\mu^*,x,a,c)\\
    = \ &   p_{a,c}(2\eta(x,a,c)-1) - (\mathbf{\lambda}^*_1 - \mathbf{\lambda}^*_2)^T \phi^{a,c}(\eta(x,a,c))\\
        & - (\mu^*_{1,c} - \mu^*_{2,c})^T \psi^{a,c}(\eta(x,a,c)).
    \end{split}
    \end{equation}
    where $\lambda=(\lambda_1,\lambda_2)\in\mathbb{R}^{2K_g}$, $\mu=[\mu_c]_{c=1}^{|\mathcal{C}|}$, $\mu_c=(\mu_{1,c},\mu_{2,c})\in\mathbb{R}^{2K_{l,c}}$. 
    Parameters $\lambda^*,\mu^*$ are determined from the convex minimization problem $(\lambda^*,\mu^*)\in \arg\min_{\lambda, \mu \geq 0} {H(\lambda, \mu)}$,
    \begin{small}
    \begin{equation}
    \label{optimal-solu}
    \begin{split}
     H(\lambda&,\mu) = \sum_{c\in\mathcal{C}} \sum_{a\in\mathcal{A}} \underset{X\sim \mathcal{P}^X_{a,c}}{\mathbb{E}}[\left(F(\lambda,\mu,X,a,c)\right)_{+}]\\
        & + \delta^{g} (\mathbf{\lambda}_1 + \mathbf{\lambda}_2)^T \mathbf{1}_{K_g}  + \delta^{l,c}\sum_{c \in \mathcal{C}} (\mu_{1,c} + \mu_{2,c})^T \mathbf{1}_{K_{l,c}},\\ 
    \end{split}
    \end{equation}
    \end{small}
    where $(\cdot)_{+}$ refers to $\max(\cdot,0)$, $\mathbf{1}_{K}$ is the ones vector with dimension $K$. The optimality of $\lambda^*$ provide global fairness guarantee and the optimality of $\mu_c^*$ provide local fairness guarantee for client $c$. (Proof see \textbf{Appendix B}.)
\end{theorem}
\textbf{Remark } This theorem also can be applied to the situation when we focus solely on local or global fairness. 
The only adjustment is to simplify problem~\eqref{var-problem} by removing unnecessary local or global fairness constraints.
%

%
Theorem~\ref{theo-1-the-optimal} establishes a explicit characterization of Bayes optimal federated fair classifier that relies on Bayes optimal score function $\eta$.
This representation can be utilized to ensure fairness at different stages of the training, including pre-processing, in-processing, and post-processing~\cite{bayes-2-zeng2024bayes}.
\M~treats the Bayes optimal federated fair classifier as the learning goal, achieving local and global fairness with minimal accuracy cost.

\subsection{Federated Post-Processing Procedure\label{method-2}}
\begin{table*}[!h]
\centering
\renewcommand{\arraystretch}{1}
\vspace{-20pt}
\caption{Comparison experimental result.}
\vspace{-5pt}
\label{experiment-result-table-DP}
\begin{threeparttable}
\resizebox{\linewidth}{!}{
\begin{tabular}{clccccccccc}
\hline
& Dataset & \multicolumn{3}{c}{Adult} & \multicolumn{3}{c}{ENEM} & \multicolumn{3}{c}{CelebA}  \\
\cmidrule(lr){3-5}\cmidrule(lr){6-8}\cmidrule(lr){9-11}
$\alpha$ & Method & Acc ($\uparrow$) & $\mathcal{M}_{DP}^{local}$ ($\downarrow$) & $\mathcal{M}_{DP}^{global}$ ($\downarrow$)    & Acc ($\uparrow$) & $\mathcal{M}_{DP}^{local}$ ($\downarrow$) & $\mathcal{M}_{DP}^{global}$ ($\downarrow$)    & Acc ($\uparrow$) & $\mathcal{M}_{DP}^{local}$ ($\downarrow$) & $\mathcal{M}_{DP}^{global}$ ($\downarrow$) \\ \hline

\multirow{8}{*}{0.5}

& FedAvg & 0.8381 & 0.1922 & 0.1759 & 0.7266 & 0.2030 & 0.1953 & 0.8934 & 0.1308 & 0.1441 \\

\cline{2-11}

& FedFB & 0.8158 & 0.1174 & 0.0767 & 0.7089 & 0.1625 & 0.0801 & 0.8705 & 0.1064 & 0.0723 \\

& FairFed & 0.8079 & 0.1416 & 0.0956 & 0.6974 & 0.1721 & 0.1002 & 0.8555 & 0.1163 & 0.0975 \\

& FCFL & 0.8167 & 0.0832 & 0.1479 & 0.6997 & 0.0900 & 0.1837 & 0.8391 & 0.0827 & 0.1373 \\

& \M\textsubscript{$g$} & \underline{\textbf {0.8218}} & \textbf {{0.0889}} & \textbf {0.0183*} & \textbf {\underline{0.7111}} & \textbf {0.0574} & \textbf {0.0141*} & \textbf {0.8698} & \textbf {0.0581} & \textbf {0.0154*} \\

& \M\textsubscript{$l$} & \textbf {{0.8252}*} & \textbf {0.0367*} & \textbf {0.0468} & \textbf {{0.7129}*} & \textbf {0.0241*} & \textbf {0.0473} & \textbf {{0.8734*}} & \textbf {{0.0196}*} & \textbf {{0.0515}} \\

& \M\textsubscript{$l \& g$} & \textbf {0.8214} & \textbf {\underline{0.0489}} & \textbf {\underline{0.0204}} & \textbf {0.7109} & \underline{\textbf {0.0281}} & \textbf {\underline{0.0151}} & \textbf {\underline{0.8710}} & \textbf {\underline{0.0312}} & \textbf {\underline{0.0295}} \\
\hline

\multirow{8}{*}{5}

& FedAvg & 0.8418 & 0.1820 & 0.1725 & 0.7268 & 0.1894 & 0.1814 & 0.8962 & 0.1394 & 0.1461 \\

\cline{2-11}

& FedFB & 0.8243 & 0.1134 & 0.0658 & 0.7116* & 0.1502 & 0.0807 & 0.8677 & 0.0856 & 0.0532 \\

& FairFed & 0.8157 & 0.1264 & 0.0971 & 0.7067 & 0.1600 & 0.1173 & 0.8516 & 0.1045 & 0.0874 \\

& FCFL & 0.8134 & 0.0673 & 0.1297 & 0.7031 & 0.0623 & 0.1073 & 0.8699 & 0.0585 & 0.0941 \\

& \M\textsubscript{$g$} & \textbf {0.8264}* & \textbf {0.0399} & \textbf {0.0104*} & \textbf {\underline{0.7107}} & \textbf {0.0548} & \textbf {0.0112*} & \textbf {\underline{0.8721}} & \textbf {0.0484} & \textbf {0.0192*} \\

& \M\textsubscript{$l$} & \textbf {0.8237} & \textbf {0.0252*} & \textbf {0.0215} & \textbf {0.7065} & \textbf {0.0133*} & \textbf {0.0423} & \textbf {0.8730*} & \textbf {0.0267*} & \textbf {0.0475} \\

& \M\textsubscript{$l \& g$} & \textbf {\underline{0.8244}} & \textbf {\underline{0.0259}} & \textbf {\underline{0.0138}} & \textbf {0.7049} & \textbf {\underline{0.0209}} & \textbf {\underline{0.0142}} & \textbf {0.8704} & \textbf {\underline{0.0279}} & \textbf {\underline{0.0242}} \\
\hline

\multirow{8}{*}{100}

& FedAvg & 0.8466 & 0.1802 & 0.1759 & 0.7279 & 0.1852 & 0.1746 & 0.8997 & 0.1352 & 0.1467 \\

\cline{2-11}

& FedFB & 0.8197 & 0.0901 & 0.0790 & 0.7047 & 0.0821 & 0.0698 & 0.8715 & 0.0736 & 0.0698 \\

& FairFed & 0.8267 & 0.0977 & 0.1086 & 0.7072 & 0.1027 & 0.0962 & 0.8617 & 0.0950 & 0.0862 \\

& FCFL & 0.8194 & 0.0613 & 0.0897 & 0.7014 & 0.0633 & 0.0871 & 0.8659 & 0.0587 & 0.0673 \\

& \M\textsubscript{$g$} & \textbf {{0.8297}}* & \textbf {0.0368} & \textbf {0.0282*} & \textbf {0.7118*} & \textbf {0.0228} & \textbf {0.0114*} & \textbf {{0.8724}*} & \textbf {0.0265} & \textbf {0.0182*} \\

& \M\textsubscript{$l$} & \textbf{\underline{0.8288}} & \textbf {0.0335*} & \textbf {0.0379} & \textbf {\underline{0.7107}} & \textbf {0.0189*} & \textbf {0.0209} & \textbf {0.8708} & \textbf {0.0175*} & \underline{\textbf {0.0213}} \\

& \M\textsubscript{$l \& g$} & \textbf {0.8283} & \textbf {\underline{0.0362}} & \textbf {\underline{0.0297}} & \textbf {0.7061} & \textbf {\underline{0.0193}} & \textbf {\underline{0.0119}} & \textbf {\underline{0.8712}} & \textbf {\underline{0.0211}} & \textbf {{0.0295}} \\
\hline

\end{tabular}
}
\begin{tablenotes}
        \item[*] \small The bold text indicates the result of \M. The best results are marked with *. The second-best results are underlined.
        \item[*] \small  All outcomes pass the significance test, with a p-value below the significance threshold of 0.05.
        \item[*] \small We use \textbf{FedAvg} as the baseline for optimal accuracy, without comparing it in terms of the accuracy-fairness trade-off.
      \end{tablenotes}
\vspace{-15pt}
\end{threeparttable}
\end{table*}
To develop model-agnostic fair FL approaches (\textbf{CH2}) while preserving data privacy, our characterization of optimal fair classifier naturally suggest a post-processing algorithm that estimates unknown Bayes optimal score $\eta$ with trained probabilistic classifier.
Therefore, \M~proposes a federated post-processing procedure to solve the optimal fair classifier, facilitating the collaborative enhancement of both local and global fairness among clients, as illustrated in Figure~\ref{model-framework}.
In this framework, client-specific parameter $\mu_c$ guarantees local fairness, while $\lambda$ serves as the global fairness guidance, according to Theorem~\ref{theo-1-the-optimal}.
\textbf{Firstly}, we investigate estimation of the optimal fair classifier in Theorem~\ref{theo-1-the-optimal} and formulate~\eqref{optimal-solu} as a bi-level problem.
For Bayes optimal score $\eta$, while it is not typically possible to identify these ground-truth Bayes optimal score function in practice, data-driven learning procedure allows us to train probabilistic classifier as empirical estimators.
We can utilize a wide range of FL approaches to obtain precisely probabilistic classifier trained using the cross-entropy loss~\cite{fedavg,personalized-tan2022towards}.
For the purpose of further mitigating the estimation error in probability scores generated by the FL model,we adopt model calibration~\cite{calibration-guo2017calibration} to group-wisely calibrate learned FL classifier.
To estimate $(\lambda^*,\mu^*)$, notice that the formula of $H(\lambda,\mu)$ involve local marginal distributions $\mathcal{P}^X_{a,c}$ and some statistics in $\phi^{a,c},\psi^{a,c}$ which are necessary for evaluating fairness.
Therefore, we propose to calibrate fairness via post-processing in a validation dataset $\mathcal{D}^{val}={\bigcup}_{c\in \mathcal{C}}\mathcal{D}^{val}_c$.
Denoting the validation data of sensitive group $a$ in client $c$ as $\mathcal{D}_{a,c}^{val}=\{ x_{i}^{a,c} \}_{i=1}^{N_{a,c}}$, the empirical estimation of~\eqref{F} can be represented as
\begin{equation}
\small
    \label{emp-optimal-solution}
    \begin{split}
    \widehat{F}(\lambda&,\mu,x,a,c)\\
    = \ &  \widehat{p}_{a,c}(2\widehat{\eta}(x,a,c)-1) - (\mathbf{\lambda}_1 - \mathbf{\lambda}_2)^T \widehat{\phi}^{a,c}(\widehat{\eta}(x,a,c))\\
        & - (\mu_{1,c} - \mu_{2,c})^T \widehat{\psi}^{a,c}(\widehat{\eta}(x,a,c)).
    \end{split}
\end{equation}
Furthermore, $\min_{\lambda, \mu \geq 0} {H(\lambda, \mu)}$ in~\eqref{optimal-solu} can be formulated as a bi-level problem with estimator $\widehat{F}$,
\begin{equation}
\small
    \label{bilevel}
    \begin{split}
    & \underset{\lambda \geq 0}{\min} \left\{ \widehat{H}(\lambda) = \sum_{c \in \mathcal{C}} \widehat{H}^c(\lambda) \right\}, \quad \widehat{H}^c (\lambda)  := \underset{\mu_c \geq 0}{\min}\widehat{H}^c (\lambda, \mu_c),\\
    \end{split}
\end{equation}
where the local fairness optimization task is 
\begin{equation}
\small
    \label{local-target}
    \begin{split}
    \widehat{H}^c (\lambda&, \mu_c) = \sum_{a\in\mathcal{A}} \frac{1}{N_{a,c}^{val}} \sum_{i=1}^{N_{a,c}^{val}} 
 \left(\widehat{F}(\lambda,\mu,x_i^{a,c},a,c)\right)_+\\
        & + \frac{\delta^{g}}{|\mathcal{C}|}(\mathbf{\lambda}_1 + \mathbf{\lambda}_2)^T \mathbf{1}_{K_g}  + \delta^{l,c}(\mu_{1,c} + \mu_{2,c})^T \mathbf{1}_{K_{l,c}}.\\ 
    \end{split}
\end{equation}
\textbf{Secondly}, \M~introduce a federated optimization algorithm tailored to solve the bi-level optimization~\eqref{bilevel}.
It is clear that $\widehat{H}(\lambda)$ and $\widehat{H}^c (\lambda, \mu_c)$ are still convex with respect to $\lambda$ and $\mu_c$ but not smooth.
Clients can locally apply subgradient descent~\cite{convex-optimization} or grid search~\cite{posthoc} to enforce local fairness.
%
However, considering that data is stored on the client side and must remain confidential, we are unable to optimize $\lambda$ directly.
%
Although it is possible to use subgradient in place of the gradient in traditional federated framework, such an approach is usually ineffective owing to the intrinsic slow convergence~\cite{fedcompo-yuan2021federated}.
To make the objective function smooth enough, we use the logarithmic exponential relaxation to replace the operator $(\cdot)_+$ in~\eqref{local-target},
\begin{equation*}
    r_\beta(x)=\frac{1}{\beta}  \log (1+\exp (\beta  x)).
\end{equation*}
Whenever $\beta \rightarrow \infty$, $r_\beta(x)$ reduces to $\max(x,0)$. We denote smooth form of $ \widehat{H}^c$ as $\widehat{H}_\beta^c$.

To solve the federated post-processing problem~\eqref{bilevel}, we introduce the projected gradient descend to approach the optimal solution.
We present the federated post-processing procedure in Algorithm 1 of \textbf{Appendix C}, along with its \textbf{efficiency}, \textbf{privacy} analysis, and additional discussion.
%
%

\section{Experiment}
In this section, we comprehensive evaluate the proposed \M~method on three publicly available real-world datasets.
Here we conduct extensive experiments to answer the following Research Questions (RQ): \textbf{RQ1}: Does \M~outperform the existing methods in effectively achieving a balance between accuracy and fairness in FL?
\textbf{RQ2}: Is \M~capable of adjusting the trade-off between accuracy and local-global fairness (sensitivity analysis)?
\textbf{RQ3}: How do client number influence the performance of \M?
\textbf{RQ4}: How about the efficiency of \M?
\subsection{Datasets and Experimental Settings}
Due to space limitations, the detailed information in this section is provided in \textbf{Appendix D.1}.

\noindent\textbf{Datasets} We consider three real-world benchmarks, \textbf{Adult}~\cite{adult-asuncion2007uci}, \textbf{ENEM}~\cite{enem-do2018instituto}, and \textbf{CelebA}~\cite{celeba-zhang2020celeba}, which are well-established for assessing fairness issues in FL~\cite{fairfed,bias-propagation-chang2023bias,postfair-duan2024post}.
%
%
%
\begin{table*}[!h]
\small
\centering
\renewcommand{\arraystretch}{1}
\vspace{-20pt}
\caption{Accuracy-Fairness Trade-off (Sensitivity Analysis).}
\vspace{-5pt}
\label{sensitive-result-table-DP}
\resizebox{\linewidth}{!}{
\begin{tabular}{lccccccccc}
\hline
 Dataset & \multicolumn{3}{c}{Adult} & \multicolumn{3}{c}{ENEM} & \multicolumn{3}{c}{CelebA}  \\
\cmidrule(lr){2-4}\cmidrule(lr){5-7}\cmidrule(lr){8-10}
 $(\delta_l,\delta_g)$ & Acc ($\uparrow$) & $\mathcal{M}_{DP}^{local}$ ($\downarrow$) & $\mathcal{M}_{DP}^{global}$ ($\downarrow$)    & Acc ($\uparrow$) & $\mathcal{M}_{DP}^{local}$ ($\downarrow$) & $\mathcal{M}_{DP}^{global}$ ($\downarrow$)    & Acc ($\uparrow$) & $\mathcal{M}_{DP}^{local}$ ($\downarrow$) & $\mathcal{M}_{DP}^{global}$ ($\downarrow$) \\ \hline

$(0.00,0.00)$ & 0.8186 & 0.0439 & 0.0035 & 0.7087 & 0.0187 & 0.0005 & 0.8607 & 0.0089 & 0.0015 \\

$(0.02,0.00)$ & 0.8209 & 0.0488 & 0.0040 & 0.7093 & 0.0243 & 0.0009 & 0.8688 & 0.0243 & 0.0039 \\

$(0.04,0.00)$ & 0.8209 & 0.0543 & 0.0062 & 0.7104 & 0.0459 & 0.0012 & 0.8701 & 0.0459 & 0.0052 \\

$(0.00,0.02)$ & 0.8235 & 0.0454 & 0.0339 & 0.7113 & 0.0192 & 0.0215 & 0.8713 & 0.0192 & 0.0215 \\

$(0.02,0.02)$ & 0.8255 & 0.0507 & 0.0366 & 0.7132 & 0.0261 & 0.0314 & 0.8732 & 0.0361 & 0.0314 \\

$(0.04,0.02)$ & 0.8243 & 0.0634 & 0.0400 & 0.7140 & 0.0492 & 0.0367 & 0.8740 & 0.0462 & 0.0367 \\

$(0.00,0.04)$ & 0.8238 & 0.0423 & 0.0352 & 0.7115 & 0.0220 & 0.0238 & 0.8715 & 0.0242 & 0.0438 \\

$(0.02,0.04)$ & 0.8252 & 0.0548 & 0.0448 & 0.7136 & 0.0332 & 0.0379 & 0.8736 & 0.0432 & 0.0479 \\

$(0.04,0.04)$ & 0.8265 & 0.0667 & 0.0516 & 0.7157 & 0.0540 & 0.0515 & 0.8757 & 0.0604 & 0.0505 \\
\hline

\end{tabular}
}
\end{table*}
\noindent \textbf{Baselines}  Since \textit{no previous work was found that endeavors to simultaneously achieve local and global fairness} within a FL framework, we compare the performance of \M~with traditional \textbf{FedAvg}~\cite{fedavg} and three SOTA methods tailored for either global or local fairness, namely \textbf{FairFed}~\cite{fairfed}, \textbf{FedFB}~\cite{fedfb-zeng2021improving}, \textbf{FCFL}~\cite{FCFL-cui2021addressing}. 
%
%
Meanwhile, we adapt \M~to focus solely on local or global fairness in FL, denoted as \M$_l$ and \M$_g$. \M$_{l\&g}$ indicates the algorithm simultaneously achieving local and global fairness.

%
%
%

%
\noindent \textbf{Evaluation Protocols} \textit{(1) Firstly}, we partition each dataset into 
a 70\% training set and the remaining 30\% for test set, while post-processing models use half of training set as validation set following previous post-processing works~\cite{postpro-pmlr-v202-xian23b,posthoc}. 
\textit{(2) Secondly}, to simulate the statistical heterogeneity in FL context, we control the heterogeneity of the sensitive attribute distribution at each client by determining the proportion of local sensitive group data based on a Dirichlet distribution $Dir(\alpha)$ as proposed by~\citet{fairfed}.
%
%
In this case, each client will possess a dominant sensitive group, and a smaller value of $\alpha$ will further reduce the data proportion of the other group, which \textit{indicates greater heterogeneity across clients}.
%
%
\textit{(3) Thirdly}, The number of participating clients is set to 5 to simulate the FL environment.
\textit{(4) fourthly}, we evaluate the FL model with Accuracy (Acc), global fairness metric $\mathcal{M}^{global}$ and maximal local fairness metric among clients $\mathcal{M}^{local}$.
Since we are interesting in DP and EO criteria, the model's fairness is assessed by local and global DP/EO metrics ($\mathcal{M}_{DP/EO}^{local},\mathcal{M}_{DP/EO}^{global}$),
smaller values of fairness metrics denote a fairer model.
\subsection{Overall Comparison (RQ1)}
We conduct extensive experiments to compare \M~with other existing fair FL baseline under varying degrees of statistical heterogeneity, and the results of DP criterion ($\mathcal{M}_{DP}$) are presented in Table~\ref{experiment-result-table-DP}. 
Here we set $\delta^{l,c}=\delta^g=0.01$.
It is essential to note that obtaining both high accuracy and fairness is challenging due to the inherent trade-off between these metrics.
All the results of EO criterion  ($\mathcal{M}_{EO}$) are reported in {\textbf{Appendix D.3}.
\noindent\textbf{Comparison result on various datasets.} From the table we can see that \M~generally demonstrates superior performance in achieving a balance between accuracy and fairness.
Compared with FedFB and FairFed, which enables to achieve global fairness, \M~gets both better accuracy and global fairness.
In terms of local fairness, \M~also surpasses FCFL in accuracy and fairness behavior.
%
This results strongly confirm the effectiveness of our approach.
\noindent\textbf{Ablation study.}
To demonstrate that \M~is also applicable to scenarios focusing exclusively on either local or global fairness, we introduced \M$_l$ and \M$_g$. The results indicate that, with a single fairness constraint, these methods typically perform better in the targeted fairness notion compared to~\M$_{l\&g}$.
For local fairness, \M$_l$ implement fairer federated model in local level with best accuracy in most cases compared to other methods. Similarly, \M$_g$ also outperforms all other methods in balancing accuracy and global fairness. Meanwhile, notice that \M$_{l\&g}$ generally provides both local and global fairness guarantee at the cost of slight accuracy degradation compared to \M$_l$ and \M$_g$.

\noindent\textbf{Impact of statistical heterogeneity.}
The most significant cause of the inconsistency between local and global fairness is statistical heterogeneity in FL. 
Table~\ref{experiment-result-table-DP} explicitly presents that the gap between local fairness and global fairness progressively narrows, as the variation in client data distributions lessens. 
Our proposed method still outperforms other approaches in most case under varying degrees of heterogeneity, further demonstrating its robustness to statistical heterogeneity in FL environments.


\subsection{Accuracy-Fairness Trade-off (RQ2)}
To investigate the capability of \M~in adjusting accuracy-fairness trade-off, we report the Acc, $\mathcal{M}^{local}_{DP}$ and $\mathcal{M}^{global}_{DP}$ under different fairness relaxation of $(\delta^l,\delta^{g})$ with $\alpha=0.5$ on all three datasets in Table~\ref{sensitive-result-table-DP}. 
Here we also set the local fairness relaxation $\delta^{l,c}$ for each client to the same value, denoted as $\delta^{l}$.
\noindent\textbf{Sensitivity analysis.}
It is important to assess the sensitivity of \M’s performance with respect to sensitivity $(\delta^l,\delta^g)$.
Table~\ref{sensitive-result-table-DP} reveals that with a constant global constraints $\delta^g$, an decreasing in $\delta^l$ results in lower Acc and local fairness, indicating that while the model achieves superior local fairness, its overall performance declines. Similarly, holding $\delta^l$ constant allows the control of global fairness through adjustments in $\delta^g$. These results confirm our claim that \M~can flexibly adjust the accuracy-fairness trade-off within FL.
%

%
%



\subsection{Scalability Concerning Client Number (RQ3)}
We examine the performance of \M~across a range of 2 to 50 clients on all three datasets using heterogeneity level $\alpha=0.5$. The results are shown in Figure~\ref{client-num}. 
\begin{figure}[t!]
  \centering
  \vspace{-10pt}
    \subfloat[Effect of client number on three datasets.]
     {\label{client-num}
    \includegraphics[width=\linewidth]{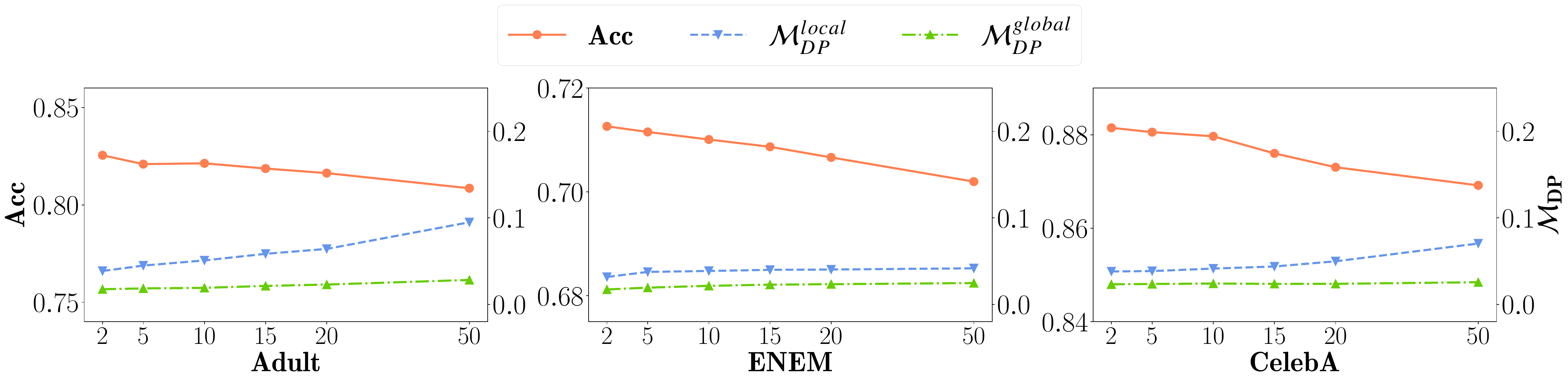}}\\
  \vspace{-10pt}
    \subfloat[Effect of communication rounds on three datasets.]
    {\label{round}
    \includegraphics[width=\linewidth]{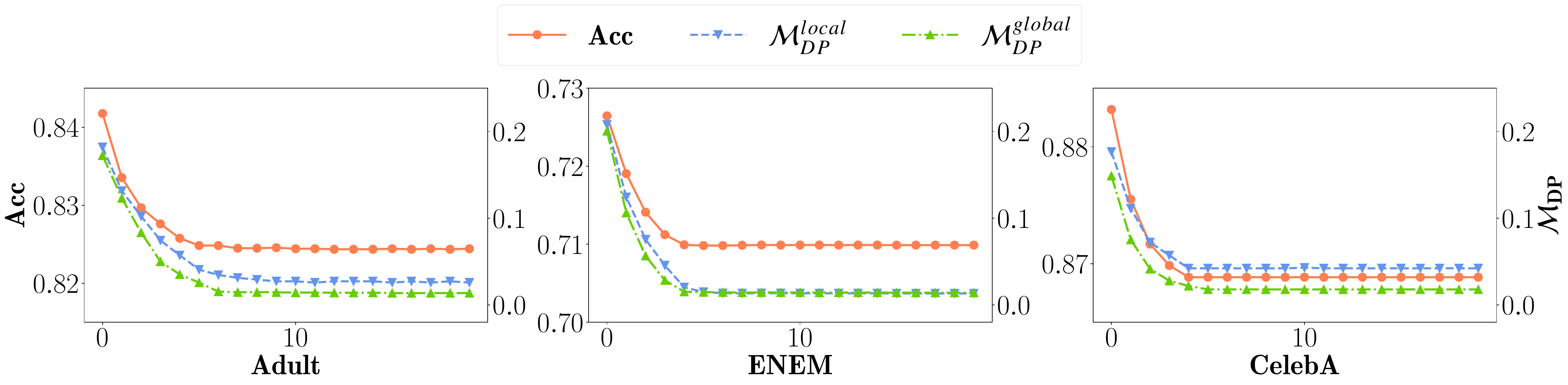}}
    \caption{Scalability and effectiveness analysis.}
      \vspace{-15pt}
\end{figure}
%

%
\noindent\textbf{Scalability.} Local fairness metric in the Adult and CelebA datasets presents a minor increase with the growing number of clients. 
This trend can be attributed to the reduction in data samples available for local fairness evaluation, which exacerbates the estimation error of federated post-processing procedure. 
Other measures reveals slight fluctuations in accuracy and fairness metrics, underscoring the model's robustness to variations in client number.
\subsection{Effectiveness Analysis (RQ4)}
In this section, we conduct out experiments to examine the communication cost introduced by~\M.

\noindent\textbf{Communication Efficiency.}
The behavior of \M~is monitored for different values of communication rounds $T$, leading to the results shown in Figure~\ref{round}.
The rapid convergence of the three metrics to stable values within 10 rounds across three datasets empirically confirms the effectiveness of~\M.


%

\section{Conclusion}
This paper proposes a novel post-processing framework for achieving \textbf{Lo}cal and \textbf{G}l\textbf{o}bal \textbf{Fair}ness within FL, namely LoGoFair. To the best of our knowledge, we are the first to offer local and global fairness in FL. The goal of \M~is to learn the Bayes optimal classifier under local and global fairness constraints in order to achieve both fairness notions with maximal accuracy. Furthermore, \M~introduces the federated post-processing procedure to solve the optimal fair classifier in a model-agnostic manner accounting for communication cost and data privacy. This approach enables participating clients to collaboratively enhance global fairness while refining their local fairness. Experiments on three publicly available real-world datasets confirm that \M~outperforms existing methods in achieving superior local and global fairness with competitive accuracy.

\section*{Acknowledgments}
This work was supported in part by the National Natural Science Foundation of China (No.62172362).

\bibliography{aaai25}
\clearpage
\appendix
\section{Discussion on Fairness Notions \label{fair-form-appendix}}
In this section, we delve into our proposed fairness notion \textit{composite linear disparity}. Firstly, we present the motivation for introducing the definition of \textit{composite linear disparity}. Secondly, we provide examples to illustrate the application of our proposed fairness notion within FL.

\subsection{Motivation}
As defined in its formulation, \textit{composite linear disparity} consists of two key components: linear disparities and the composite maximum of these disparities.
Concerning linear disparities, given that group fairness seeks to ensure that ML models deliver equitable treatment to individuals across different sensitive attributes, most fairness notions are defined by measuring the disparity in the treatment of sensitive groups. 
These concepts generally concentrate on the disparity in the positive prediction rate between specific paired groups, determined by sensitive attributes.
Previous work~\cite{bayes-2-zeng2024bayes} has indicated that many significant fairness notions can be represented by a linear function of the Bayes optimal score such as DP~\cite{DP-dwork2012fairness} and EOP~\cite{EO-hardt2016equality}.
%
Examples are given in next subsection to verify this claim.
By incorporating the linear disparity notion to characterize fairness, we extend the applicability of our study to a class of fairness notions. 
%

Concerning composite maximum, in terms of fairness metrics within multiple paired groups, existing works~\cite{beyond-alghamdi2022beyond,postpro-pmlr-v202-xian23b,posthoc,fair-guarantees-denis2024fairness} often measure fairness by the maximum disparity among such paired groups, e.g. EO~\cite{EO-hardt2016equality}, which goes beyond the representation capacity of linear disparity.
To address the this limitations, \textit{composite linear disparity} considers the composite maximum of these linear disparities as formulated in its definition in Section 4.
By integrating the composite maximum with linear disparity, the \textit{composite linear disparity} concept encompasses a wide range of established fairness notions, and is effective in characterizing fairness in multi-group situations.
%

%
In the FL context, \textit{composite linear disparity} has several advantages compared to other fairness definitions, especially when considering both local and global group fairness: (1) \textbf{Simplicity in Representation.} Directly extending existing fairness notions to the FL setting can result in complex formulaic representations~\cite{fairfed}. \textit{Composite linear disparity} simplifies the expression of fairness notions within FL.
(2) \textbf{Flexibility.} Given that our fairness concept covers various fairness definitions, it enables the implementation of different fairness notions at the local and global levels in FL, and even allows for achieving multiple non-conflicting fairness notions at one side through the composite maximum.

Subsequently, we will give some examples to demonstrate the representation of common fairness notions by composite linear disparity, in centralized settings or within FL, to highlight its advantages in the context of FL fairness evaluation.
\subsection{Examples}
In this section, we primarily discuss DP and EO metrics characterized by \textit{composite linear disparity} in the centralized setting and FL settings to demonstrate the applicability
of our proposed fairness notion.

\noindent\textbf{In the centralized setting}, we complete the proof of Example~\ref{exmple-1}.
\noindent\textbf{Proof of Example~\ref{exmple-1}.} We have a sensitive attribute $a \in \mathcal{A} = \{-1, 1\}$ and $\widehat{Y}=h(X,A)$.\\
(1) For DP criterion, 
\begin{equation*}
    \small
    \begin{split}
         \mathcal{M}_{DP}(h)&=\left | P(\widehat{Y} = 1 | A = -1) - P(\widehat{Y} = 1 | A = 1) \right |\\
         &= \left|\sum_{a \in \mathcal{A}} aP(\widehat{Y} = 1 | A = a) \right|\\
         &= \left|\sum_{a \in \mathcal{A}} a \underset{X \sim \mathcal{P}^X_{a}}{\mathbb{E}}\left[ h(X,a) \right] \right|\\
    \end{split}
\end{equation*}
Hence, let $K=1$, $\mathcal{M}_{DP}(h)=\left| \mathscr{D}_1 (h) \right|$, and $\phi_1^a=a$, we get DP metric.\\
(2) For EO criterion, 
\begin{equation*}
    \small
    \begin{split}
        \mathcal{M}_{EO}(h) & =\underset{y \in \{0,1\}}{\max}\left | P(\widehat{Y}  = 1 | A = -1, Y = y) - \right.\\
         & \qquad\qquad\qquad\qquad\qquad\left. P(\widehat{Y} = 1 | A = 1, Y = y) \right |.\\
         & = \underset{y \in \{0,1\}}{\max}\left | \sum_{a \in \mathcal{A}}aP(\widehat{Y}  = 1 | A = a, Y = y)\right |\\
         & = \underset{y \in \{0,1\}}{\max}\left | \sum_{a \in \mathcal{A}}a \frac{P(\widehat{Y}  = 1, Y = y | A = a)P(A=a)}{P( A=a,Y = y)}\right |\\
         & = \underset{y \in \{0,1\}}{\max}\left | \sum_{a \in \mathcal{A}}a \frac{P(\widehat{Y}  = 1, Y = y | A = a)}{P(Y = y|A=a)}\right |\\
         & := \underset{k=1,2}{\max}\left | \mathscr{D}_k(h)\right |,\\
    \end{split}
\end{equation*}
while
\begin{equation*}
    \small
    \begin{split}
    P(\widehat{Y}  = 1, Y = 1 | A = a) &= \underset{X \sim \mathcal{P}^X_{a}}{\mathbb{E}}\left[ \eta(X,a)h(X,a) \right],\\
    P(\widehat{Y}  = 1, Y = 0 | A = a) &= \underset{X \sim \mathcal{P}^X_{a}}{\mathbb{E}}\left[ (1-\eta(X,a))h(X,a) \right].\\
    \end{split}
\end{equation*}
We have
\begin{equation*}
\small
    \begin{split}
    \mathscr{D}_1 (h) &=\sum_{a\in \mathcal{A}} \underset{X \sim \mathcal{P}^X_{a}}{\mathbb{E}}\left[\frac{ a\eta(X,a)}{P(Y=1|A=a)}h(X,a)\right],\\
    \mathscr{D}_2 (h) &=\sum_{a\in \mathcal{A}} \underset{X \sim \mathcal{P}^X_{a}}{\mathbb{E}}\left[\frac{ a(1-\eta(X,a))}{P(Y=0|A=a)}h(X,a)\right].\\
    \end{split}
\end{equation*}
Hence, let $K=2$ and $\phi_1^a(w)=aw/P(Y=1\mid A=a), \phi_2^a(w)=a(1-w)/P(Y=0 \mid A=a)$, we get EO metric.
\noindent\textbf{In the FL setting}, we provide formula of local/global DP and EO metrics represented by \textit{composite linear disparity}. Following pervious work~\cite{fairfed,AGLFOP-hamman2024demystifying}, local fairness metrics defined on the individual data distribution of each client $P(X,A,Y|C)$ and global fairness metrics defined on global distribution $P(X,A,Y)$. We first recall our local and global fairness defined by \textit{composite linear disparity}:
\begin{equation*}
\small
    \begin{split}
    {\rm(Local) \ }&: \quad \mathcal{M}^{l,c}_{\psi} (h)=\max_{k_{l}=1,\ldots,K_{l,c}} \left| \mathscr{D}^{l,c}_{k_{l}} (h) \right|, \\
    \mathscr{D}^{l,c}_{k_{l}}(h)&=\sum_{a \in \mathcal{A}} \underset{X\sim \mathcal{P}^X_{a,c}}{\mathbb{E}}\left[ \psi^{a,c}_{k_{l}}( \eta(X,a,c) ) h(X,a,c) \right],\\
    {\rm(Global)}&: \quad \mathcal{M}^g_{\phi} (h)=\max_{k_g=1,\ldots,K_g} \left| \mathscr{D}^g_{k_g} (h) \right|,\\
    \mathscr{D}^g_{k_g}(h)&=\sum_{a \in \mathcal{A}} \sum_{c\in\mathcal{C}} \underset{X\sim \mathcal{P}^X_{a,c}}{\mathbb{E}}\left[ \phi_{k_g}^{a,c} ( \eta(X,a,c) ) h(X,a,c) \right],
    \end{split}
\end{equation*}
\noindent \textbf{DP metric:} Local DP fairness for client $c$ and global DP fairness are defined as
\begin{equation*}
    \small
    \begin{split}
        \mathcal{M}_{DP}^{l,c}(h)&=\bigg| P(\widehat{Y} = 1 | A = -1,C=c)\\
         &\qquad\qquad\qquad\qquad-P(\widehat{Y} = 1 | A = 1,C=c) \bigg|,\\
         \mathcal{M}_{DP}^{g}(h)&=\left | P(\widehat{Y} = 1 | A = -1) -P(\widehat{Y} = 1 | A = 1) \right|\\
         &=\left| \sum_{a\in\mathcal{A}}\sum_{c\in\mathcal{C}} a P(\widehat{Y} = 1 | A = a,C=c)P(C=c|A=a)\right|.\\
        \end{split}
    \end{equation*}
We can formulate local and global DP fairness notions by \textit{composite linear disparity}:
\begin{equation*}
    \small
    \begin{split}
        \mathcal{M}_{DP}^{l,c}(h)&=\left|\sum_{a \in \mathcal{A}} a \underset{X \sim \mathcal{P}^X_{a,c}}{\mathbb{E}}\left[ h(X,a,c) \right] \right|,\\
        \mathcal{M}_{DP}^{g}(h) &=\left | \sum_{a\in\mathcal{A}} \sum_{c\in\mathcal{C}} \underset{X \sim \mathcal{P}^X_{a,c}}{\mathbb{E}}\left[a P(C=c|A=a) h(X,a,c) \right] \right|.\\
        \end{split}
    \end{equation*}
Hence, let $K_{l,c}=1$, $\mathcal{M}_{DP}^{l,c}(h)=\left| \mathscr{D}_1^{l,c} (h) \right|$, and $\psi^{a,c}_{1}=a$, we get local DP metric for client $c$.
Let $K_{g}=1$, $\mathcal{M}_{DP}^{g}(h)=\left| \mathscr{D}_1 (h) \right|$, and $\phi^{a,c}_{1}=a P(C=c|A=a)$, we get global DP metric.
\noindent\textbf{EO metric:} Local EO fairness for client $c$ and global EO fairness are defined as
\begin{equation*}
    \small
    \begin{split}
    \small
         \mathcal{M}_{EO}^{l,c}&=\underset{y \in \{0,1\}}{\max}\left | P(\widehat{Y}  = 1 | A = -1, Y = y,C=c) \right.\\
         &\qquad\qquad\quad- \left. P(\widehat{Y} = 1 | A = 1, Y = y,C=c) \right |.\\
         \mathcal{M}_{EO}^{g}&=\underset{y \in \{0,1\}}{\max}\left | P(\widehat{Y}  = 1 | A = -1, Y = y) \right.\\
         &\qquad\qquad\quad\qquad\quad- \left. P(\widehat{Y} = 1 | A = 1, Y = y) \right |\\
        &=\underset{y \in \{0,1\}}{\max}\bigg\lvert \sum_{a\in\mathcal{A}}\sum_{c\in\mathcal{C}} aP(\widehat{Y}  = 1 | A = a, Y = y,C=c)\\
         &\qquad \qquad \qquad \qquad \qquad \quad  \times P(C=c \mid A=a,Y=y) \bigg\rvert.\\
    \end{split}
\end{equation*}
We can formulate local and global EO fairness notions by \textit{composite linear disparity}:
In terms of local EO fairness for client $c$,
\begin{equation*}
    \small
    \begin{split}
    \small
         \mathcal{M}_{EO}^{l,c}&=\underset{y \in \{0,1\}}{\max}\left | P(\widehat{Y}  = 1 | A = -1, Y = y,C=c) \right.\\
         &\qquad\qquad\quad- \left. P(\widehat{Y} = 1 | A = 1, Y = y,C=c) \right |\\
         & =\underset{y \in \{0,1\}}{\max}\left| \sum_{a\in\mathcal{A}}a\frac{P(\widehat{Y} = 1,Y = y | A = 1,C=c)}{P(Y = y |A = 1,C=c)} \right |\\
         & := \underset{k_l=1,2}{\max}\left | \mathscr{D}_{k_{l}}^{l,c}(h)\right |,\\
    \end{split}
\end{equation*}
where 
\begin{equation*}
\small
    \begin{split}
    \mathscr{D}_1^{l,c} (h) &=\sum_{a\in \mathcal{A}} \underset{X \sim \mathcal{P}^X_{a,c}}{\mathbb{E}}\left[\frac{ a\eta(X,a,c)}{P(Y=1|A=a,C=c)}h(X,a,c)\right],\\
    \mathscr{D}_2^{l,c} (h) &=\sum_{a\in \mathcal{A}} \underset{X \sim \mathcal{P}^X_{a,c}}{\mathbb{E}}\left[\frac{ a(1-\eta(X,a,c))}{P(Y=0|A=a,C=c)}h(X,a,c)\right].\\
    \end{split}
\end{equation*}
In terms of global EO fairness,
\begin{equation*}
    \small
    \begin{split}
    \small
         \mathcal{M}_{EO}^{g}&=\underset{y \in \{0,1\}}{\max}\left |\sum_{a\in\mathcal{A}}\sum_{c\in\mathcal{C}} aP(\widehat{Y}  = 1 | A = a, Y = y,C=c) \right.\\
         &\qquad \qquad \qquad \qquad \qquad \quad \left. \times P(C=c \mid A=a,Y=y) \right|\\
         & = \underset{y \in \{0,1\}}{\max}\left |\sum_{a\in\mathcal{A}}\sum_{c\in\mathcal{C}} \frac{a}{P(A=a,Y=y)}P(A=a,C=c)\right.\\
         &\qquad \qquad \qquad \quad  \left. \times P(\widehat{Y}=1,Y=y \mid A=a,C=c) \right|\\
        & := \underset{k_g=1,2}{\max}\left | \mathscr{D}_{k_{g}}^{g}(h)\right |,\\
    \end{split}
\end{equation*}
where, with $p_{a,c}:=P(A=c,C=c)$,
\begin{equation*}
\small
    \begin{split}
    \mathscr{D}_1^{g} (h) &=\sum_{a\in \mathcal{A}} \underset{X \sim \mathcal{P}^X_{a,c}}{\mathbb{E}}\left[\frac{ ap_{a,c}\eta(X,a,c)}{P(A=a,Y=1)}h(X,a,c)\right],\\
    \mathscr{D}_2^{g} (h) &=\sum_{a\in \mathcal{A}} \underset{X \sim \mathcal{P}^X_{a,c}}{\mathbb{E}}\left[\frac{ ap_{a,c}(1-\eta(X,a,c))}{{P(A=a,Y=0)}}h(X,a,c)\right].\\
    \end{split}
\end{equation*}
Hence, let $K_{l,c}=2$, $\mathcal{M}_{EO}^{l,c}(h)=\underset{k_l=1,2}{\max} \left| \mathscr{D}_{k_l}^{l,c} (h) \right|$, and $\psi_1^{a,c}={ a\eta(X,a,c)}/{P(Y=1|A=a,C=c)}$, $\psi_2^{a,c}={ a(1-\eta(X,a,c))}/{P(Y=0|A=a,C=c)}$, we get local EO metric.
Let $K_{g}=2$, $\mathcal{M}_{EO}^{g}(h)=\underset{k_g=1,2}{\max} \left| \mathscr{D}_{k_g}^{g} (h) \right|$, and $\phi_1^{a,c}={ ap_{a,c}\eta(X,a,c)}/{P(A=a,Y=1)}$, $\phi_2^{a,c}={ap_{a,c}(2-d\eta(X,a,c))}/{P(A=a,Y=1)}$, we get global EO metric.
\clearpage

\section{Proof of Theorem~\ref{theo-1-the-optimal} \label{proofs}}
We list several Lemmas here.
\begin{lemma}~\cite{lemma-subgradient-Bertsekas1973StochasticOP}
\label{subgradient-lemma}
    The subdifferential of the function $F(x)=E\{f(x, \omega)\}$ at a point $x$ is given by
\begin{equation*}
\partial F(x)=E\{\partial f(x, \omega)\}
\end{equation*} 
where $f(\cdot,\omega)$ is a real-value convex function and the set $E\{\partial f(x, \omega)\}$ is defined as
\begin{equation*}
    \begin{split}
        E\{\partial f(x, &\omega)\} = \int_{\Omega} \partial f(x, \omega) d P(\omega) \\
        &= \bigg\{x^* \in \mathbb{R}^n \mid x^*=\int_{\Omega} x^*(\omega) d P(\omega), \\
        &\quad x^*(\cdot): \text{measurable}, x^*(\omega) \in \partial f(x, \omega) \text{ a.e.} \bigg\}.
    \end{split}
\end{equation*}     
\end{lemma}
\begin{lemma}~\cite{Variational-Analysis-RockWets98}
\label{maximum}
Let $f_1, \dots, f_m:$ $\mathbb{R}^n \rightarrow (-\infty, +\infty]$ be convex functions. Define $f(x) = \max \{ f_1(x), \dots, f_m(x) \}, \quad \forall x \in \mathbb{R}^n$.
For \( x_0 \in \bigcap_{i=1}^m \text{dom} f_i \), define \( I(x_0) = \{ i \mid f_i(x_0) = f(x_0) \} \). Then
\[
\partial f(x_0) = \text{conv} \bigcup_{i \in I(x_0)} \partial f_i(x_0).
\]
\end{lemma}
\begin{lemma}~\cite{Variational-Analysis-RockWets98}
\label{optimal-lemma}
    Let $f:\mathbb{R}^d \to \mathbb{R}$ be a convex continuous function. We consider the minimizer $x^*$ of the function $f$ over the set $B$. Then for $x^*$ to be locally optimal it is necessary that 
    \begin{equation*}
        \partial f({x^*})+\mathcal{N}_B({x^*}) \ni 0,
    \end{equation*}
    where $\mathcal{N}_B$ denotes the normal cone of set $B$.
    If $B=\mathbb{R}^d_+$, let $\mathcal{K}:=\left\{k \in[d], x_k^* \neq 0\right\}$. Then there exists a subgradient $\xi \in \partial f\left(x^*\right)$, such that for all $k \in[d]$ we have $\xi_k \geq 0$ and $\forall k \in\mathcal{K}, \xi_k=0$. 
\end{lemma}

We present the proofs of Theorem~\ref{theo-1-the-optimal} in this section.\\
\noindent\textbf{Proof sketch} Initially, we investigate the Lagrangian function to obtain the form dual of Problem~\eqref{var-problem}, leading to a classifier described by~\eqref{F} and~\eqref{optimal-solu}. We then demonstrate the feasibility and optimality of this classifier, thereby verifying that it is the Bayes optimal federated fair classifier.

\noindent\textbf{1) Solution of form dual problem.}
We firstly represent the problem~\eqref{var-problem} in a detailed form. Start from exploring the risk $\mathcal{R}(h)$ over federated data distribution $(X,A,Y,C)$, where $X \in \mathcal{X} \subset \mathbb{R}^d$, $A \in \mathcal{A}= \{ -1,1 \}$, $Y \in \mathcal{Y}= \{ 0,1 \}$, and $C \in \mathcal{C}$. 
Let $p_{a,c}:=P(A=a,C=c)$,
\begin{equation*}
    \small
    \begin{split}
    \mathcal{R}&(h) = \mathbb{P}(h(X, A, C) \neq Y)\\
    & =\mathbb{P}(h(X, A, C)=0, Y=1)\\
    & \quad  +\mathbb{P}(h(X, A, C)=1, Y=0)\\
    & =\mathbb{P}(Y=1)-\mathbb{P}(h(X, A, C)=1, Y=1) \\
    & \quad +\mathbb{P}(h(X, A, C)=1)-\mathbb{P}(h(X, A, C)=1, Y=1)\\
    & =\mathbb{P}(Y=1)+\sum_{a \in \mathcal{A}}\sum_{ c \in \mathcal{C}} p_{a,c} \underset{X\sim \mathcal{P}^X_{a,c}}{\mathbb{E}}[h(X, A, C)]\\
    & \quad- 2 \sum_{a \in \mathcal{A}}\sum_{ c \in \mathcal{C}} p_{a,c}\underset{X \sim \mathcal{P}^X_{a,c}}{\mathbb{E}}[h(X,a,c) \eta(X,a,c) ]\\
    & =\mathbb{P}(Y=1)\\
    & \quad - \sum_{a \in \mathcal{A}}\sum_{ c \in \mathcal{C}} p_{a,c}\underset{X\sim \mathcal{P}^X_{a,c}}{\mathbb{E}} [h(X,a,c)(2\eta(X,a,c)-1)] \\
    \end{split}
\end{equation*}
Denoting $\mathscr{D}^{l,c}(h)=[\mathscr{D}^{l,c}_{k_l}(h)]_{k_l=1}^{K_{l,c}}$ and $\mathscr{D}^g(h)=[\mathscr{D}^g_{k_g}(h)]_{k_g=1}^{K_g}$, problem~\eqref{var-problem} can be formulated as
\begin{equation}
{\small
    \label{var-problem-1}}
    \begin{split}
     &\min _{h \in \mathcal{H}} \mathcal{R}(h),\\
     \text { s.t. } & |\mathscr{D}^{l,c}(h)| \leq \delta^{l,c}, c \in [C],\\
     & |\mathscr{D}^{g}_{k_g}(h)| \leq \delta^{g}. \\
    \end{split}
\end{equation}
where the constraints represent the absolute value of every element within the vector less than given fairness levels.
To this problem, the first step is to construct the Lagrangian function. 
For $\lambda=(\lambda_1,\lambda_2)\in\mathbb{R}^{2K_g}$, $\mu=[\mu_c]_{c=1}^{|\mathcal{C}|}$, $\mu_c=(\mu_{1,c},\mu_{2,c})\in\mathbb{R}^{2K_{l,c}}$, let $\lambda_i=(\lambda_i^{(1)},\ldots, \lambda_i^{(K_g)}) \in \mathbb{R}^{K_g}$ and $\mu_{i,c}=(\mu_{i,c}^{(1)},\ldots,\mu_{i,c}^{(K_{l,c})}) \in \mathbb{R}^{K_{l,c}}$, $c \in C$, $i\in\{1,2\}$.
%
%
The Lagrangian function  $L(h, \lambda, \mu)$
\begin{equation*}
\small
    \begin{split}
    &L(h, \lambda, \mu)=\mathcal{R}(h)\\
        & + \sum_{k_g=1}^{K_g} (\mathbf{\lambda}_1^{(k_g)} - \mathbf{\lambda}_2^{(k_g)}) \sum_{a\in\mathcal{A}}\sum_{c\in\mathcal{C}} \underset{X\sim \mathcal{P}^X_{a,c}}{\mathbb{E}}[\phi_{k_g}^{a,c}(\eta(X,a,c)) h(X,a,c)]\\
        & +\sum_{c=1}^C \sum_{k_l=1}^{K_{l,c}}(\mu_{1,c}^{(k_l)} - \mu_{2,c}^{(k_l)}) \sum_{a\in\mathcal{A}}\underset{X\sim \mathcal{P}^X_{a,c}}{\mathbb{E}}[\psi_{k_l}^{a,c}(\eta(X,a,c)) h(X,a,c)] \\
        & \qquad- \delta^{g} \sum_{k_g=1}^{K_g} (\mathbf{\lambda}_1^{(k_g)} + \mathbf{\lambda}_2^{(k_g)} ) - \delta^{l,c}\sum_{c=1}^C \sum_{k_l=1}^{K_{l,c}} (\mu_{1,c}^{(k_l)} + \mu_{2,c}^{(k_l)}).\\ 
         &= \mathcal{R}(h) + (\mathbf{\lambda}_1 - \mathbf{\lambda}_2)^T \mathscr{D}^{g}(h) +\sum_{c=1}^C (\mu_{1,c} - \mu_{2,c})^T \mathscr{D}^{l,c}(h)\\
        & \qquad- \delta^{g} (\mathbf{\lambda}_1 + \mathbf{\lambda}_2)^T \mathbf{1}_{K_g}  - \sum_{c=1}^C \delta^{l,c} (\mu_{1,c} + \mu_{2,c})^T \mathbf{1}_{K_{l,c}},\\ 
    \end{split}
\end{equation*}
where $\mathbf{1}_{K}$ is the ones vector with dimension $K$.
The form dual of problem~\eqref{var-problem} can be written as 
\begin{equation*}
\max_{\lambda,\mu \geq 0} \min _{h \in \mathcal{H}} L(h, \lambda, \mu).
\end{equation*}
Denoting $\phi^{a,c}:=[ \phi^{a,c}_{k_g} ]_{k_g=1}^{K_g}$ and $\psi^{a,c}:=[ \phi^{a,c}_{k_l} ]_{k_l=1}^{K_{l,c}}$, intergrating our transform in risk $\mathcal{R}(h)$, we can derive that 
\begin{align*} 
\small
& L(h, \lambda, \mu)\\
& =\mathbb{P}(Y=1) \\
& - \sum_{a\in\mathcal{A}}\sum_{c\in\mathcal{C}} \underset{X\sim \mathcal{P}^X_{a,c}}{\mathbb{E}} [p_{a,c}h(X,a,c)(2\eta(X,a,c)-1)]\\
& +  \sum_{a\in\mathcal{A}}\sum_{c\in\mathcal{C}} \underset{X\sim \mathcal{P}^X_{a,c}}{\mathbb{E}}[ (\lambda_1-\lambda_2)^T \phi^{a,c}(\eta(X,a,c))h(X, a, c)]\\
&+ \sum_{a\in\mathcal{A}}\sum_{c\in\mathcal{C}} \underset{X\sim \mathcal{P}^X_{a,c}}{\mathbb{E}}[ (\mu_{1,c} -\mu_{2,c} )^T \psi^{a,c}(\eta(X,a,c))h(X, a, c)]\\
& - \delta^{g} (\mathbf{\lambda}_1 + \mathbf{\lambda}_2)^T \mathbf{1}_{K_g}  - \sum_{c=1}^C \delta^{l,c} (\mu_{1,c} + \mu_{2,c})^T \mathbf{1}_{K_{l,c}}\\
& = \mathbb{P}(Y=1) - \sum_{a\in\mathcal{A}}\sum_{c\in\mathcal{C}} \underset{X\sim \mathcal{P}^X_{a,c}}{\mathbb{E}} [F(\lambda,\mu,X,a,c)h(X,a,c)]\\
& \qquad - \delta^{g} (\mathbf{\lambda}_1 + \mathbf{\lambda}_2)^T \mathbf{1}_{K_g}  - \sum_{c=1}^C \delta^{l,c} (\mu_{1,c} + \mu_{2,c})^T \mathbf{1}_{K_{l,c}},\\
\end{align*}
where 
\begin{align*} 
\small
& F(\lambda,\mu,X,a,c) \\
& = p_{a,c}(2\eta(X,a,c)-1) - (\lambda_1-\lambda_2)^T \phi^{a,c}(\eta(X,a,c)) \\
& \qquad - (\mu_{1,c} -\mu_{2,c} )^T \psi^{a,c}(\eta(X,a,c)).
\end{align*}

The inner optimization problem requires one to minimize the Lagrangian $L(h, \lambda, \mu)$ with respect to $h$. It is explicit that a solution is to take $h^*(X,a,c)=\mathbb{I}\left[F(\lambda,\mu,X,a,c) \geq 0\right]$.
Here, since $h^*$ is determined by parameters $\lambda,\mu$, we denote it by $h^*_{\lambda,\mu}(x,a,c)$.
%
Plug the classifier $h^*_{\lambda,\mu}$ into the outer maximization optimization, we have
\begin{equation*}
\small
    \begin{split}
        L(h^*_{\lambda,\mu}&, \lambda, \mu) \\ 
        = & \mathbb{P}(Y=1) - \sum_{a\in\mathcal{A},c\in\mathcal{C}}  \underset{X\sim \mathcal{P}^X_{a,c}}{\mathbb{E}}[\left(F(\lambda,\mu,X,a,c)\right)_{+}]\\
        & - \delta^{g} (\mathbf{\lambda}_1 + \mathbf{\lambda}_2)^T \mathbf{1}_{K_g}  - \delta^{l,c}\sum_{c \in \mathcal{C}} (\mu_{1,c} + \mu_{2,c})^T \mathbf{1}_{K_{l,c}}.\\ 
    \end{split}
\end{equation*}
Here $(\cdot)_{+}$ presents $\max(\cdot,0)$. The next step is to solve the outer optimization problem $\max_{\lambda,\mu \geq 0} L(h^*_{\lambda,\mu}, \lambda, \mu).$
Let $H: \mathbb{R}^{2(K_g+\sum_c K_{l,c})}_+ \to \mathbb{R}$ be the function
\begin{equation*}
\small
    \begin{split}
        & H(\lambda,\mu) = \sum_{a\in\mathcal{A}}\sum_{c\in\mathcal{C}}  \underset{X\sim \mathcal{P}^X_{a,c}}{\mathbb{E}}[\left(F(\lambda,\mu,X,a,c)\right)_{+}]\\
        & \qquad + \delta^{g} (\mathbf{\lambda}_1 + \mathbf{\lambda}_2)^T \mathbf{1}_{K_g}  + \delta^{l,c}\sum_{c \in \mathcal{C}} (\mu_{1,c} + \mu_{2,c})^T \mathbf{1}_{K_{l,c}}.\\ 
    \end{split}
\end{equation*}
It is clear that maximize $ L(h^*_{\lambda,\mu}, \lambda, \mu)$ with respect to $\lambda,\mu \geq 0$ is equivalent to $\min_{\lambda,\mu\geq0}H(\lambda,\mu)$. So far we derived characterization in Theorem~\ref{theo-1-the-optimal}.
The rest of the proof consists in showing that such a solution $h^*_{\lambda^*,\mu^*}(X,a,c)$, obtained by optimizing parameter $\lambda,\mu$ in $\min_{\lambda,\mu\geq0}H(\lambda,\mu)$, implies that $h^*_{\lambda^*,\mu^*}(X,a,c)$ is indeed local and global fair with minimal misclassificaition risk.
\noindent\textbf{2) Feasibility and Optimality.}
First, we prove the optimality of $\lambda^*$ provide global fairness guarantee. 
Since $F(\lambda,\mu,X,a,c)$ is linear to $\lambda$ and $\mu$, it is obvious that $H(\lambda,\mu)$ is convex and continuous in $\lambda$ and $\mu$.
We derive the subgradient $\xi_{\lambda_{i}^*}^{(k_g)}$ of $H$ at the optimum ${\lambda^*_{i}}^{(k_g)},i=1,2, k_g\in [K_g]$. With Lemma~\ref{subgradient-lemma},
\begin{equation*}
\small
    \begin{split}
        &\xi_{\lambda_{1}^*}^{(k_g)} = \partial_{{\lambda_1^*}^{(k_g)}}H(\lambda,\mu)\\
        &\quad = \sum_{a\in\mathcal{A}}\sum_{c\in\mathcal{C}}  \underset{X\sim \mathcal{P}^X_{a,c}}{\mathbb{E}}[\partial_{{\lambda_1^*}^{(k_g)}}\left(F(\lambda,\mu,X,a,c)\right)_{+}] + \delta^{g}\\
        & \quad = \sum_{a\in\mathcal{A}}\sum_{c\in\mathcal{C}} \left\{ \underset{X\sim \mathcal{P}^X_{a,c}}{\mathbb{E}}\left[\phi^{a,c}_{k_g}(\eta(X,a,c))\left(\mathbb{I}[F(\lambda,\mu,X,a,c) > 0] \right.\right. \right.\\
        & \qquad\qquad \left. + u_{1}^{(k_g)}\mathbb{I}[F(\lambda,\mu,X,a,c)=0])\right] \Bigg\} + \delta^{g}\\
    \end{split}
\end{equation*}
where the second equation derives from Lemma~\ref{subgradient-lemma}, and the third equation is from Lemma~\ref{maximum}, $u_{1}^{(k_g)}\in [0,1]$ for all $k_g$.
Under Assumption~\ref{assumpt-1}, since the Bayes optimal score has no atom and $F$ is linear to $\lambda$, $F(\lambda,\mu,X,a,c)=0$ only hold at a null set.
Consequently, plugging in $h^*_{\lambda,\mu}(X,a,c)=\mathbb{I}\left[F(\lambda,\mu,X,a,c) \geq 0\right]$, we have 
\begin{equation*}
\small
    \begin{split}
        \xi_{\lambda_{1}^*}^{(k_g)} =& \partial_{{\lambda_1^*}^{(k_g)}}H(\lambda,\mu)\\
        = &-\sum_{a\in\mathcal{A}}\sum_{c\in\mathcal{C}}\underset{X\sim \mathcal{P}^X_{a,c}}{\mathbb{E}}\left[\phi^{a,c}_{k_g}(\eta(X,a,c))h^*_{\lambda,\mu}(X,a,c)\right]+ \delta^{g}.\\
        = & -\mathscr{D}^{g}_{k_g}(h) + \delta^{g}
    \end{split}
\end{equation*}
Similarly, for $\xi_{\lambda_{2}^*}^{(k_g)}$, we have
\begin{equation*}
\small
    \begin{split}
        \xi_{\lambda_{2}^*}^{(k_g)} =& \partial_{{\lambda_2^*}^{(k_g)}}H(\lambda,\mu)\\
        = &\sum_{a\in\mathcal{A}}\sum_{c\in\mathcal{C}}\underset{X\sim \mathcal{P}^X_{a,c}}{\mathbb{E}}\left[\phi^{a,c}_{k_g}(\eta(X,a,c))h^*_{\lambda,\mu}(X,a,c)\right]+ \delta^{g}.\\
        = & \mathscr{D}^{g}_{k_g}(h) + \delta^{g}
    \end{split}
\end{equation*}
Considering paired parameter ${\lambda^*_{i}}^{(k_g)},i=1,2$, by Lemma~\ref{optimal-lemma}, if ${\lambda^*_{1}}^{(k_g)}>0,{\lambda^*_{2}}^{(k_g)}>0$, we have 
\begin{equation*}
\small
   \mathscr{D}^{g}_{k_g}(h) = -\delta^{g},
   \mathscr{D}^{g}_{k_g}(h) = \delta^{g},
\end{equation*}
which leads to a contradiction.
If ${\lambda^*_{1}}^{(k_g)}=0,{\lambda^*_{2}}^{(k_g)}=0$, we have 
\begin{equation*}
\small
   \mathscr{D}^{g}_{k_g}(h) \geq -\delta^{g},
   \mathscr{D}^{g}_{k_g}(h) \leq \delta^{g}.
\end{equation*}
If ${\lambda^*_{1}}^{(k_g)}=0,{\lambda^*_{2}}^{(k_g)}>0$, we have
\begin{equation*}
\small
   \mathscr{D}^{g}_{k_g}(h) \geq -\delta^{g},
   \mathscr{D}^{g}_{k_g}(h) = \delta^{g}.
\end{equation*}
If ${\lambda^*_{1}}^{(k_g)}>0,{\lambda^*_{2}}^{(k_g)}=0$, we have 
\begin{equation*}
\small
   \mathscr{D}^{g}_{k_g}(h) = -\delta^{g},
   \mathscr{D}^{g}_{k_g}(h) \leq \delta^{g}.
\end{equation*}
Overall, we have shown that for all $k_g \in [K_g]$, $|\mathscr{D}^{g}_{k_g}(h^*_{\lambda^*,\mu})| \leq \delta^{g}$.

Secondly, the optimality of $\mu^*$ provide local fairness guarantee. The proof techniques are extremely similar to our proof with respect to $\lambda^*$.
Here, we omit the proof related to $\mu^*$.
Lastly, it is necessary to prove that the classifier $h^*_{\lambda^*,\mu^*}$ can achieve minimum misclassification risk among classifiers satisfying local and global fairness constraints.
From the proof above, we get that
\begin{equation*}
\small
\begin{split}
    0 =& ({\lambda^*_{1}}^{(k_g)}) \xi_{\lambda_{1}^*}^{(k_g)} + ({\lambda^*_{2}}^{(k_g)}) \xi_{\lambda_{2}^*}^{(k_g)} \\
    =& ({\lambda^*_{1}}^{(k_g)}) (-\mathscr{D}^{g}_{k_g}(h) + \delta^{g}) + ({\lambda^*_{2}}^{(k_g)}) (\mathscr{D}^{g}_{k_g}(h) + \delta^{g})\\
    =& -({\lambda^*_{1}}^{(k_g)} - {\lambda^*_{2}}^{(k_g)}) \mathscr{D}^{g}_{k_g}(h)  + ({\lambda^*_{1}}^{(k_g)} + {\lambda^*_{2}}^{(k_g)}) \delta^{g}.
\end{split}
\end{equation*}
Generally, we get that
\begin{equation*}\small
\begin{split}
    -(\mathbf{\lambda}_1 - \mathbf{\lambda}_2)^T \mathscr{D}^{g}(h) + \delta^{g} (\mathbf{\lambda}_1 + \mathbf{\lambda}_2)^T \mathbf{1}_{K_g} = 0.
\end{split}
\end{equation*}
For $\mu^*$, the similar corollary also holds true,
\begin{equation*}
\small
\begin{split}
    -\sum_{c=1}^C (\mu_{1,c} - \mu_{2,c})^T \mathscr{D}^{l,c}(h) + \sum_{c=1}^C \delta^{l,c} (\mu_{1,c} + \mu_{2,c})^T \mathbf{1}_{K_{l,c}} = 0.
\end{split}
\end{equation*}
Consequently, the Lagrangian function equals to risk function when plugging in optimal classifier, $L(h^*_{\lambda^*,\mu^*}, \lambda^*, \mu^*)=\mathcal{R}(h^*_{\lambda^*,\mu^*})$.
For other classifiers $h$ under constraints, since these classifiers satisfy fairness constraints, we have $ L(h.\lambda^*_h,\mu^*_h) \leq \mathcal{R}(h)$, where $\lambda^*_h,\mu^*_h$ is the optimal parameters for classifier $h$.
Hence,
\begin{equation*}
\small
\begin{split}
    \mathcal{R}(h^*_{\lambda^*,\mu^*}) = L(h^*_{\lambda^*,\mu^*}, \lambda^*, \mu^*) \leq L(h, \lambda^*_h, \mu^*_h) \leq \mathcal{R}(h).
\end{split}
\end{equation*}
We finish the proof.

\subsection{Example for Theorem~\ref{theo-1-the-optimal} \label{proofs-example}}

For the DP metric, we have shown its representation under the proposed fairness notion in Example 1. 
Plugging this representation into equations~\eqref{F} and~\eqref{optimal-solu} in Theorem~\ref{theo-1-the-optimal}, we obtain the explicit characterization of the federated optimal fair classifier under local and global DP fairness constraints:
The calibration function 
\begin{equation*}
\begin{split}
F(&\lambda^*,\mu^*,x,a,c)\\
&=\pi_{a,c} (2\eta(X_i,a,c)-1) - (\lambda_1^* - \lambda_2^*)\frac{\pi_{a,c}}{p_{a}} - (\mu_{1,c}^* - \mu_{2,c}^*).
\end{split}
\end{equation*}
Thus, we have the federated optimal fair classifier 
\begin{equation*}
\begin{split}
h^*(&x,a,c)\\
&= \mathbb{I} \left[\eta(X_i,a,c) \geq\frac{1}{2} + (\lambda_1^* - \lambda_2^*)\frac{1}{p_{a}}+(\mu_{1,c}^* - \mu_{2,c}^*)\frac{1}{\pi_{a,c}}\right].
\end{split}
\end{equation*}
The parameters $\lambda^*,\mu^*$ can be obtained by solving the convex problem in~\eqref{optimal-solu}.
The Bayesian optimal classifier for the EO metric can also be derived using a similar approach.

\section{Algorithm and Analysis \label{algorithm-appendix}}
\subsection{Algorithm}
We provide the detailed algorithm of \M~in Algorithm~\ref{algorithm-total}.
\begin{algorithm}[h!]
\caption{LoGoFair} 
\label{algorithm-total}
\textbf{Input}: validation dataset $\mathcal{D}_c^{val}$ from client $c \in \mathcal{C}$;
communication round $T$; local interval S; calibrated pre-trained probabilistic classifier $\widehat{\eta}$; learning rate $\{\gamma_g,\gamma_l \}$; initial parameters $\lambda^{(0)}=(\lambda^{(0)}_1,\lambda^{(0)}_2), \mu^{(0)}=[\mu_c^{(0)}]_{c=1}^{|\mathcal{C}|}$\\
\textbf{Output:} federated fair classifier $\hat{h}$ 
%
\begin{algorithmic} 
\STATE characterize Bayes optimal federated fair classifier $h^*(x,a,c)$ with $H(\lambda,\mu)$, and formulate local objectives $\widehat{H}_\beta^c(\lambda,\mu)$
\STATE $t=0$, $\lambda^{(0)}=\mu^{(0)}=0, c\in\mathcal{C}$
\WHILE{$t < T$}
\FOR{\textbf{client} $c$ \textbf{in parallel} }
\STATE set $\lambda^{(t)}_{c,0} = \lambda^{(t)}, \mu_{c,0}^{(t)}=\mu_{c}^{(t)}$
\FOR{$s=0,1,\ldots,S-1$}
\STATE update $\mu_{c,s+1}^{(t)}= \left(\mu_{c,s}^{(t)} + \gamma_l \nabla_{\mu} \widehat{H}_\beta^c (\lambda^{(t)}, \mu_{c,s}^{(t)}) \right)_+$
\STATE update  $\lambda_{c,s+1}^{(t)} = \left( \lambda_{c,s}^{(t)} + \gamma_g \nabla_{\lambda} \widehat{H}_\beta^c (\lambda_{c,s}^{(t)}, \mu_{c,s}^{(t)}) \right)_+$
\ENDFOR
\STATE send $\Delta\lambda_{c}^{(t)}=\lambda_{c,S}^{(t)}-\lambda^{(t)}$ to server,set $\mu_c^{(t+1)}=\mu_{c,S}^{(t)}$
\ENDFOR
\STATE \textbf{server side}:
\STATE  update $\lambda^{(t+1)}=\lambda^{(t)}+\sum_{c\in\mathcal{C}}\Delta \lambda^{(t)}_c$, send it to clients 
\STATE $t++$
\ENDWHILE
\STATE solve local problem in~\eqref{bilevel} for $\hat{h}(x,a,c)$ 
\end{algorithmic}
\end{algorithm}
\subsection{Privacy and Efficiency Analysis of~\M}
In this section, we evaluate the proposed \M~framework in terms of privacy concerns, computation efficiency, and communication efficiency.
\noindent\textbf{Privacy.} 
\M~ensures data privacy in both stages. \textit{In the first stage}, inferring the Bayes optimal federated fair classifier requires clients to provide local statistics related to $\phi$. 
For example, as presented in Appendix~\ref{fair-form-appendix}, to evaluate global DP, the statistic ${N_a}$ is necessary to estimate $P(C=c|A=a)$ for client $c$, where $N_a$ denotes the total sample size of group $a$.
These statistics are the least amount of information necessary to evaluate global fairness~\cite{fairfed}.
The server can not deduce extra distribution information of data or sensitive groups in addition to sample size.
We can even employ encryption federated aggregation such as SecAgg~\cite{SecAgg-10.1145/3133956.3133982} to prevent any form of information leakage.
\textit{In the second stage}, \M~only entails sharing global fairness guidance $\lambda$. Since this parameter does not enable external inference of the local classifier, our algorithm maintains data privacy in the second stage as well.
\noindent\textbf{Computation efficiency.} The primary computation cost incurred by \M~is the iterative optimization of $\lambda$ and $\mu_c$. 
In most cases, $\lambda$ and $\mu_c$ are low-dimensional vectors.
For instance, regrading DP, as depicted in Appendix~\ref{fair-form-appendix}, both $\lambda$ and $\mu_c$ are two-dimensional; regrading EO, both $\lambda$ and $\mu_c$ are four-dimensional.
Consequently, our algorithm exhibits significantly low computational overhead, with a local iteration time complexity of $\mathcal{O}(N_c)$.
\noindent\textbf{Communication efficiency.} Our proposed method is communication-efficient. In terms of \textit{transmission cost}, we only need to transmit a low-dimensional vector $\lambda$. Regarding \textit{communication rounds}, we prove that the local fairness optimization objective $\widehat{H}_\beta^c(\lambda,\mu_c)$ is \textit{convex} and \textit{L-smooth} with respect to $\lambda$ and $\mu_c$. Existing work~\cite{fedcompo-yuan2021federated} has shown that, under these conditions, the convergence rate of such federated optimization algorithm can reach $\mathcal{O}(\frac{1}{T})$.

\begin{theorem}
    $\widehat{H}_\beta^c(\lambda,\mu_c)$ is convex and L-smooth.
\end{theorem}
\noindent\textit{Proof.} Convexity is trivial here. Since $\widehat{F}$ is linear to $\lambda,\mu$ and $(\cdot)_+$ operator preserves convexity, $\widehat{H}_\beta^c(\lambda,\mu_c)$ is convex to $\lambda,\mu$.
Next, we aim to prove L-smooth. Since $\widehat{H}_\beta^c(\lambda,\mu_c)$ is evidently second-order differentiable, our initial task is to calculate the Hessian matrix. Denoting $\widetilde{\phi}^{a,c}_i = \left(-{\phi}^{a,c}(\eta(x_i,a_i,c)),{\phi}^{a,c}(\eta(x_i,a_i,c))\right)$, the Hessian matrix can be writen as
\begin{equation*}
    \begin{split}
        \nabla_\lambda^2 \widehat{H}^c_{\beta}\left(\lambda, \mu_c\right)=\sum_{a \in \mathcal{A}} \frac{1}{N_{a, c}^{v a l}} \sum_{i=1}^{N_{a, c}^{v a l}} r_\beta^{\prime \prime}(\widehat{F})(\widetilde{\phi}^{a, c}_i)(\widetilde{\phi}^{a, c}_i)^T,
    \end{split}
\end{equation*}
where
\begin{equation*}
    \begin{split}
        r_\beta^{\prime \prime}(x)=\frac{\beta \exp (\beta x)}{(1+\exp (\beta x))^2} \leq \frac{\beta}{4}.
    \end{split}
\end{equation*}
Consider the maximum eigenvalue $\epsilon_{\max }$ of the Hessian matrix. It can be expressed as
\begin{equation*}
    \begin{split}
        \epsilon_{\max }=\max _{\|\mathbf{v}\|=1} \mathbf{v}^T \nabla_\lambda^2 \widehat{H}_\beta^c\left(\lambda, \mu_c\right) \mathbf{v}.
    \end{split}
\end{equation*}
Substituting the Hessian matrix, we have,
\begin{equation*}
    \begin{split}
        \epsilon_{\max }=\max _{\|\mathbf{v}\|=1} \sum_{a \in \mathcal{A}} \frac{1}{N_{a, c}^{\mathrm{val}}} \sum_{i=1}^{N_{a, c}^{\mathrm{val}}} r_\beta^{\prime \prime}(\widehat{F})\left(\mathbf{v}^T \widetilde{\phi}_i^{a, c}\right)^2.
    \end{split}
\end{equation*}
Combining $ r_\beta^{\prime \prime}(x) \leq \frac{\beta}{4}$, the upper bound of the maximum eigenvalue of the Hessian matrix can be written as:
\begin{equation*}
    \begin{split}
        \epsilon_{\max } \leq \frac{\beta}{4} \max _{a \in \mathcal{A}} \max _{i \leq N_{a, c}^{\text {val }}}\left\|\widetilde{\phi}_i^{a, c}\right\|^2
    \end{split}
\end{equation*}
Hence, $\widehat{H}_\beta^c(\lambda,\mu_c)$ is L-smooth to $\lambda$. Consider $\nabla_{\mu_c}^2 \widehat{H}^c_{\beta}\left(\lambda, \mu_c\right)$, we can get same result for $\mu_c$.

\subsection{Discussion on AGLFOP and LoGoFair}
\citet{AGLFOP-hamman2024demystifying} introduced AGLFOP to investigate the theoretical bounds of the accuracy-fairness trade-off, aiming to identify the optimal performance achievable given global data distributions and prediction outcomes.
In this section, we discuss the relationships and differences between AGLFOP and LoGoFair.
The primary purpose of both AGLFOP and LoGoFair is to explore the theoretical optimality of accuracy-fairness trade-offs in FL.
The main difference lies in the solution methodology: AGLFOP proves the framework’s convexity and solves it using convex solvers, thereby defining the theoretical limits of accuracy-fairness trade-offs. LoGoFair derives the explicit formulation of the optimal fair classifier and introduces a practical post-processing procedure based on this result. Thus, AGLFOP identifies the best performance that any FL strategy can attain, while LoGoFair provides a state-of-the-art algorithm to achieve the optimal trade-offs.
\section{Additional Experimental Details and Results \label{experiment-appendix}}
\subsection{Experimental Details}
\subsubsection{Datasets}
\begin{itemize}
    \item The \textbf{Adult} dataset~\cite{adult-asuncion2007uci} comprises more than 45000 samples based on 1994 U.S. census data, where the task is to predict whether the annual income of an individual is above \$50,000. We consider the gender of each individual as the sensitive attribute and train the logistic regression as the classification model.
    \item The \textbf{ENEM} dataset~\cite{enem-do2018instituto} contains about 1.4 million samples from Brazilian college entrance exam scores along with student demographic information. We follow~\cite{beyond-alghamdi2022beyond} to quantized the exam score into 2 classes as label, and consider race as sensitive attribute. We train multilayer perceptron (MLP) as the classification model.
    \item The \textbf{CelebA} dataset~\cite{celeba-zhang2020celeba} is a facial image dataset consists of about 200k instances with 40 binary attribute annotations. We identify the binary feature \textit{smile} as target attributes which aims to predict whether the individuals in the images exhibit a smiling expression. The \textit{race} of individuals is chosen as sensitive attribute. We train Resnet18~\cite{resnet-he2016deep} on CelebA as as the classification model. 
\end{itemize}

The determination of sensitive attributes and labels on three datasets has been verified significant in previous research~\cite{beyond-alghamdi2022beyond,benchmark-han2024ffb}.

\subsubsection{Baselines}
Since \textit{no previous work that endeavors to simultaneously achieve local and global fairness} within a FL framework was found, we compare the performance of \M~with traditional \textbf{FedAvg}~\cite{fedavg} and three SOTA methods tailored for either global or local fairness, namely \textbf{FairFed}~\cite{fairfed}, \textbf{FedFB}~\cite{fedfb-zeng2021improving}, \textbf{FCFL}~\cite{FCFL-cui2021addressing}. 
\begin{itemize}
    \item \textbf{FedAvg} serves as a core Federated Learning model and provides the baseline for our experiments. It works by computing updates on each client’s local dataset and subsequently aggregating these updates on a central server via averaging.
    \item \textbf{FairFed} introduces an approach to adaptively adjust the aggregation weights of different clients based on their local fairness metric to train federated model with global fairness guarantee. 
    \item \textbf{FedFB} presents a FairBatch-based approach~\cite{fairbatch} to compute the coefficients of FairBatch parameters on the server. This method integrates global reweighting for each client into the FedAvg framework to fulfill fairness objectives.
    \item \textbf{FCFL} proposed a two-stage optimization to solve a multi-objective optimization with fairness constraints. The prediction loss at each local client is treated as an objective, and FCFL maximize the worst-performing client while considering fairness constraints by optimizing a surrogate maximum function involving all objectives.
\end{itemize}
Meanwhile, we adapt \M~to focus solely on local or global fairness in FL, denoted as \M$_l$ and \M$_g$. \M$_{l\&g}$ indicates the algorithm simultaneously achieving local and global fairness.

\noindent \textbf{Evaluation Protocols} \textit{(1) Firstly}, we partition each dataset into 
a 70\% training set and the remaining 30\% for test set, while post-processing models use half of training set as validation set following previous post-processing works~\cite{postpro-pmlr-v202-xian23b,posthoc}. Our post-processing method is implemented on classifier pre-trained by FedAvg.
\textit{(2) Secondly}, to simulate the statistical heterogeneity in FL context, we control the heterogeneity of the sensitive attribute distribution at each client by determining the proportion of local sensitive group data based on a Dirichlet distribution $Dir(\alpha)$ as proposed in~\cite{fairfed}.
In this case, each client will possess a dominant sensitive group, and a smaller value of $\alpha$ will further reduce the data proportion of the other group, which \textit{indicates greater heterogeneity across clients}.
%
%
\textit{(3) Thirdly}, The number of participating clients is set to 5 to simulate the FL environment.
\textit{(4) fourthly}, we evaluate the FL model with Accuracy (Acc), global fairness metric $\mathcal{M}^{global}$ and maximal local fairness metric among clients $\mathcal{M}^{local}$.

\subsubsection{Parameter Settings} We provide hyperparameter selection ranges for each model in Table~\ref{tab:hyperparameters}.
For all other hyperparameters, we follow the codes provided by authors and retain their default parameter settings.
\begin{table}[hb!]
\small
\centering
\caption{Hyperparameter Selection Ranges}
\label{tab:hyperparameters}
\begin{tabular}{llp{3.5cm}}
\toprule
\textbf{Model} & \textbf{Hyperparameter} & \textbf{Ranges} \\
\midrule
\multirow{6}{*}{\textbf{General}} 
& Learning rate & \{0.001, 0.005, 0.01\} \\
& Global round & \{20, 30, 50, 80\} \\
& Local round & \{5, 10, 20, 30\} \\
& Local batch size & \{128, 256, 512\} \\
& Hidden layer & \{16, 32, 64\} \\
& Optimizer & \{Adam, Sgd\} \\
\midrule
\textbf{FedFB} & Step size (\(\alpha\)) & \{0.005, 0.01, 0.05, 0.3\} \\
\midrule
\multirow{2}{*}{\textbf{FairFed}} 
& Fairness budget (\(\beta\)) & \{0.01, 0.05, 0.5, 1\} \\
& Local debiasing (\(\alpha\)) & \{0.005, 0.01, 0.05\} \\
\midrule
{\textbf{FCFL}} 
& Fairness constraint (\(\epsilon\)) & \{0.01, 0.03, 0.05, 0.07\} \\
\bottomrule
\end{tabular}
\end{table}
\subsubsection{Experiments Compute Resources} We conducted our experiments on a GPU server equipped with 8 CPUs and an NVIDIA RTX 4090 (24G).
\begin{table*}[b]
\centering
\renewcommand{\arraystretch}{1}
\vspace{-20pt}
\caption{Comparison experimental result (EO).}
\vspace{-5pt}
\label{experiment-result-table-EO}
\begin{threeparttable}
\resizebox{\linewidth}{!}{
\begin{tabular}{clccccccccc}
\hline
& Dataset & \multicolumn{3}{c}{Adult} & \multicolumn{3}{c}{ENEM} & \multicolumn{3}{c}{CelebA}  \\
\cmidrule(lr){3-5}\cmidrule(lr){6-8}\cmidrule(lr){9-11}
$\alpha$ & Method & Acc ($\uparrow$) & $\mathcal{M}_{EO}^{local}$ ($\downarrow$) & $\mathcal{M}_{EO}^{global}$ ($\downarrow$)    & Acc ($\uparrow$) & $\mathcal{M}_{EO}^{local}$ ($\downarrow$) & $\mathcal{M}_{EO}^{global}$ ($\downarrow$)    & Acc ($\uparrow$) & $\mathcal{M}_{EO}^{local}$ ($\downarrow$) & $\mathcal{M}_{EO}^{global}$ ($\downarrow$) \\ \hline

\multirow{8}{*}{0.5}

& FedAvg & 0.8381 & 0.2249 & 0.1410 & 0.7266 & 0.2437 & 0.2741 & 0.8934 & 0.0815 & 0.0696 \\

\cline{2-11}

& FedFB & 0.8242 & 0.1750 & 0.0977 & 0.7112 & 0.1973 & 0.1058 & 0.8823 & 0.0648 & 0.0427 \\

& FairFed & 0.8125 & 0.1416 & 0.1048 & 0.7019 & 0.1805 & 0.1771 & 0.8478 & 0.0741 & 0.0590 \\

& FCFL & 0.8291 & 0.0848 & 0.1195 & 0.7054 & 0.1011 & 0.1730 & 0.8779 & 0.0453 & 0.0519 \\

& \M\textsubscript{$g$} & \textbf{{0.8350}*} & \textbf {{0.0961}} & \textbf {0.0293*} & \textbf {{0.7157}*} & \textbf {0.0678} & \textbf {0.0289*} & \textbf {0.8872}* & \textbf {0.0485} & \textbf {0.0253*} \\

& \M\textsubscript{$l$} & \textbf {{0.8302}} & \textbf {0.0465*} & \textbf {0.0665} & \textbf {\underline{0.7139}} & \textbf {0.0328*} & \textbf {0.0515} & \textbf {\underline{0.8851}} & \textbf {{0.0276}*} & \textbf {{0.0430}} \\

& \M\textsubscript{$l \& g$} & \textbf {\underline {0.8334}} & \textbf {\underline{0.0529}} & \textbf {\underline{0.0346}} & \textbf {0.7092} & \underline{\textbf {0.0345}} & \textbf {\underline{0.0310}} & \textbf {{0.8845}} & \textbf {\underline{0.0312}} & \textbf {\underline{0.0285}} \\
\hline

\multirow{8}{*}{5}

& FedAvg & 0.8429 & 0.1899 & 0.1346 & 0.7268 & 0.2261 & 0.2584 & 0.8962 & 0.0783 & 0.0554 \\

\cline{2-11}

& FedFB & 0.8318 & 0.1249 & 0.0721 & \underline{0.7156} & 0.1525 & 0.0870 & 0.8843 & 0.0586 & 0.0332 \\

& FairFed & 0.8145 & 0.1172 & 0.0893 & 0.7051 & 0.1837 & 0.1261 & 0.8696 & 0.0733 & 0.0642 \\

& FCFL & 0.8280 & 0.0824 & 0.1015 & 0.7097 & 0.0879 & 0.1123 & 0.8817 & 0.0423 & 0.0518 \\

& \M\textsubscript{$g$} & \textbf {0.8364}* & \textbf {0.0615} & \textbf {0.0270*} & \textbf {{0.7164}*} & \textbf {0.0351} & \textbf {0.0246*} & \textbf {{0.8879}*} & \textbf {0.0393} & \textbf {0.0221*} \\

& \M\textsubscript{$l$} & \textbf {0.8312} & \textbf {0.0413*} & \textbf {0.0576} & \textbf {0.7141} & \textbf {0.0285*} & \textbf {0.0408} & \textbf {0.8858} & \textbf {0.0258*} & \textbf {0.0364} \\

& \M\textsubscript{$l \& g$} & \textbf {\underline{0.8340}} & \textbf {\underline{0.0487}} & \textbf {\underline{0.0296}} & \textbf {0.7120} & \textbf {\underline{0.0309}} & \textbf {\underline{0.0287}} & \textbf {\underline{0.8862}} & \textbf {\underline{0.0297}} & \textbf {\underline{0.0240}} \\
\hline

\multirow{8}{*}{100}

& FedAvg & 0.8466 & 0.1807 & 0.1285 & 0.7279 & 0.2145 & 0.2390 & 0.8997 & 0.0759 & 0.0476 \\

\cline{2-11}

& FedFB & 0.8357 & 0.0824 & 0.0590 & \underline{0.7162} & 0.0818 & 0.0614 & 0.8869 & 0.0417 & 0.0308 \\

& FairFed & 0.8203 & 0.0977 & 0.1086 & 0.7091 & 0.1375 & 0.1180 & 0.8790 & 0.0526 & 0.0497 \\

& FCFL & 0.8294 & 0.0714 & 0.0832 & 0.7104 & 0.0646 & 0.0735 & 0.8832 & 0.0370 & 0.0402 \\

& \M\textsubscript{$g$} & \textbf {{0.8378}}* & \textbf {0.0409} & \textbf {0.0241*} & \textbf {0.7187*} & \textbf {0.0301} & \textbf {0.0236*} & \textbf {{0.8895}*} & \textbf {0.0298} & \textbf {0.0213*} \\

& \M\textsubscript{$l$} & \textbf{\underline{0.8352}} & \textbf {0.0322*} & \textbf {0.0347} & \textbf {{0.7138}} & \textbf {0.0277*} & \textbf {0.0357} & \textbf {\underline{0.8880}} & \textbf {0.0236*} & {\textbf {0.0293}} \\

& \M\textsubscript{$l \& g$} & \textbf {0.8335} & \textbf {\underline{0.0397}} & \textbf {\underline{0.0262}} & \textbf {0.7142} & \textbf {\underline{0.0280}} & \textbf {\underline{0.0249}} & \textbf {{0.8874}} & \textbf {\underline{0.0277}} & \textbf {\underline{0.0225}} \\
\hline

\end{tabular}
}
\begin{tablenotes}
        \item[*] \small The bold text indicates the result of \M. The best results are marked with *. The second-best results are underlined.
        \item[*] \small  All outcomes pass the significance test, with a p-value below the significance threshold of 0.05.
        \item[*] \small We use \textbf{FedAvg} as the baseline for optimal accuracy, without comparing it in terms of the accuracy-fairness trade-off.
      \end{tablenotes}
\end{threeparttable}
\end{table*}
\subsection{Experimental Results of EO}
\subsubsection{Overall Comparison (RQ1)}
To compare \M~with other existing fair FL baseline under varying degrees of statistical heterogeneity, the experimental results of EO criterion ($\mathcal{M}_{EO}$) are presented in Table~\ref{experiment-result-table-EO}. 
Here we set $\delta^{l,c}=\delta^g=0.02$.

\noindent\textit{Comparison result.} 
We note a slight decline in the performance of all methods when the EO fairness criterion is applied. This is attributed to the fact that EO involves a finer group partitioning than DP, leading to smaller sample sizes per group and thus diminishing the model's generalization capability.
From the table we can see that \M~generally demonstrates superior performance in achieving a balance between accuracy and fairness.
\M~gets both better accuracy and global fairness compared with FedFB and FairFed.
In terms of local fairness, \M~also surpasses FCFL in accuracy and fairness behavior.
%
This results confirm our claim in the experiments section and empirically verify the effectiveness of~\M.
\begin{table*}[t]
\small
\centering
\renewcommand{\arraystretch}{1}
\vspace{-20pt}
\caption{Accuracy-Fairness Trade-off for EO (Sensitivity Analysis).}
\vspace{-5pt}
\label{sensitive-result-table-EO}
\resizebox{\linewidth}{!}{
\begin{tabular}{lccccccccc}
\hline
 Dataset & \multicolumn{3}{c}{Adult} & \multicolumn{3}{c}{ENEM} & \multicolumn{3}{c}{CelebA}  \\
\cmidrule(lr){2-4}\cmidrule(lr){5-7}\cmidrule(lr){8-10}
 $(\delta_l,\delta_g)$ & Acc ($\uparrow$) & $\mathcal{M}_{EO}^{local}$ ($\downarrow$) & $\mathcal{M}_{EO}^{global}$ ($\downarrow$)    & Acc ($\uparrow$) & $\mathcal{M}_{EO}^{local}$ ($\downarrow$) & $\mathcal{M}_{EO}^{global}$ ($\downarrow$)    & Acc ($\uparrow$) & $\mathcal{M}_{EO}^{local}$ ($\downarrow$) & $\mathcal{M}_{EO}^{global}$ ($\downarrow$) \\ \hline

$(0.00,0.00)$ & 0.8039 & 0.0206 & 0.0058 & 0.7027 & 0.0133 & 0.0029 & 0.8720 & 0.0121 & 0.0007 \\

$(0.02,0.00)$ & 0.8163 & 0.0447 & 0.0086 & 0.7059 & 0.0291 & 0.0059 & 0.8768 & 0.0283 & 0.0019 \\

$(0.04,0.00)$ & 0.8282 & 0.0560 & 0.0118 & 0.7090 & 0.0480 & 0.0092 & 0.8801 & 0.0474 & 0.0045 \\

$(0.00,0.02)$ & 0.8205 & 0.0239 & 0.0313 & 0.7061 & 0.0203 & 0.0284 & 0.8793 & 0.0156 & 0.0249 \\

$(0.02,0.02)$ & 0.8334 & 0.0529 & 0.0346 & 0.7092 & 0.0345 & 0.0310 & 0.8845 & 0.0312 & 0.0285 \\

$(0.04,0.02)$ & 0.8352 & 0.0634 & 0.0402 & 0.7140 & 0.0492 & 0.0344 & 0.8874 & 0.0476 & 0.0320 \\

$(0.00,0.04)$ & 0.8320 & 0.0297 & 0.0458 & 0.7103 & 0.0238 & 0.0315 & 0.8831 & 0.0218 & 0.0347 \\

$(0.02,0.04)$ & 0.8349 & 0.0362 & 0.0501 & 0.7136 & 0.0356 & 0.0427 & 0.8864 & 0.0395 & 0.0410 \\

$(0.04,0.04)$ & 0.8367 & 0.0687 & 0.0543 & 0.7175 & 0.0527 & 0.0506 & 0.8893 & 0.0516 & 0.0532 \\
\hline

\end{tabular}
}
\end{table*}
%

%
\noindent\textit{Ablation study.} 
Table~\ref{experiment-result-table-EO} demonstrates that~\M$_{l}$ and~\M$_{g}$ typically outperform~\M$_{l\&g}$ in the targeted fairness notion when only a single fairness constraint is applied.
This phenomenon also confirms our model's ability to handle either local or global fairness issues.

%
\noindent\textit{Impact of statistical heterogeneity.} 
As depicted in Table~\ref{experiment-result-table-EO}, the gap between local fairness and global fairness gradually decreases as client data distribution variation diminishes. Our proposed method consistently outperforms other approaches across most scenarios, even with different levels of heterogeneity, highlighting its robustness to statistical heterogeneity in FL environments.

\subsubsection{Accuracy-Fairness Trade-off (RQ2)}

To investigate the capability of \M~in adjusting accuracy-fairness trade-off, we report the Acc, $\mathcal{M}^{local}_{EO}$ and $\mathcal{M}^{global}_{EO}$ under different fairness relaxation of $(\delta^l,\delta^{g})$ with $\alpha=0.5$ on all three datasets in Table~\ref{sensitive-result-table-EO}. 
%
%

%
\noindent\textit{Sensitivity analysis.}
Table~\ref{sensitive-result-table-EO} reveals that through the control of one of the parameter $(\delta^l,\delta^g)$, the other can be tuned to achieve flexible trade-off adjustments.
These results confirm our claim that \M~can flexibly adjust the accuracy-fairness trade-off within FL.

\subsubsection{Scalability and Effectiveness Analysis (RQ3,RQ4)}
We examine the performance of \M~across a range of 2 to 50 clients on all three datasets using heterogeneity level $\alpha=0.5$. The results are shown in Figure~\ref{client-num-EO}. 
\begin{figure}[t!]
  \centering
  \vspace{10pt}
    \subfloat[Effect of client number on three datasets.]
     {\label{client-num-EO}
    \includegraphics[width=\linewidth]{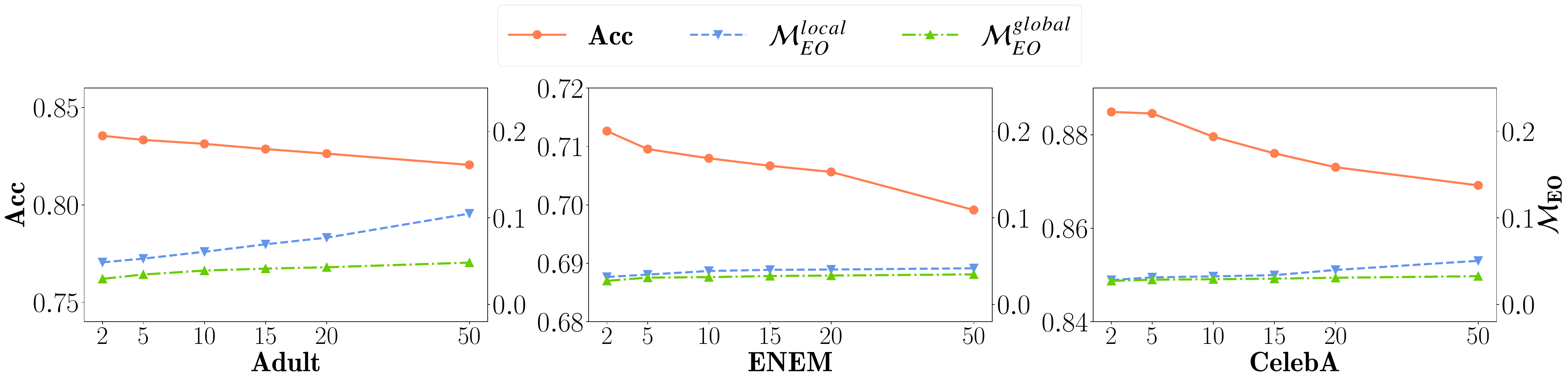}}\\
  \vspace{-1pt}
    \subfloat[Effect of communication rounds on three datasets.]
    {\label{round-EO}
    \includegraphics[width=\linewidth]{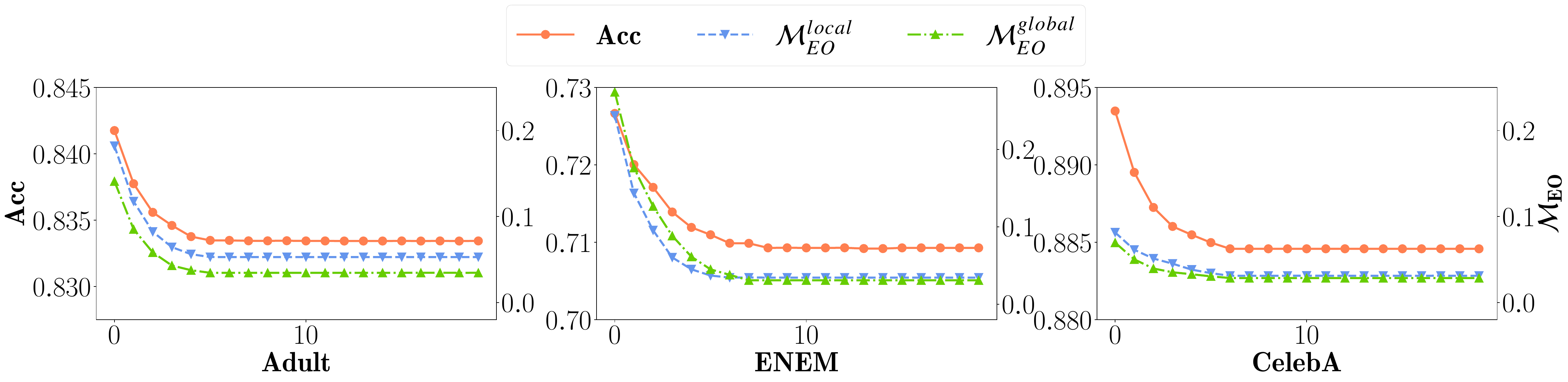}}
    \caption{Scalability and effectiveness analysis (EO).}
      \vspace{1pt}
\end{figure}

\noindent\textit{Scalability.} Similar to DP metric, the local fairness metric in the Adult and CelebA datasets exhibits a slight increase as the number of clients rises. This can be attributed to the reduction in data samples available for local fairness evaluation, which amplifies the estimation error in the federated post-processing procedure. Meanwhile, other metrics show minor fluctuations in both accuracy and fairness, underscoring the model's robustness to changes in client numbers.

\noindent\textit{Communication Efficiency.}
The behavior of \M~is monitored for different values of communication rounds $T$, leading to the results shown in Figure~\ref{round}.
The rapid convergence of the three metrics to stable values within 10 rounds across all three datasets empirically validates the effectiveness of \M.

\end{document}